\documentclass[letterpaper, 10 pt, conference]{ieeeconf}  

\IEEEoverridecommandlockouts                              

\usepackage{graphicx}
\usepackage{subfig}
\usepackage{float}
\usepackage{xcolor}
\usepackage{multirow}
\title{\LARGE \bf
ThermalDiffusion: Visual-to-Thermal Image-to-Image Translation for Autonomous Navigation
}

\author{Shruti Bansal$^{1}$, Wenshan Wang$^{1}$, Yifei Liu$^{1}$, Parv Maheshwari$^{1}$
\thanks{$^{1}$ The authors are with the
Robotics Institute, Carnegie Mellon University, Pittsburgh, PA 15213, USA. {\tt\small \{shrutib, wenshanw, yifeil5, parvm\}@andrew.cmu.edu}}%
}

\begin{document}

\maketitle
\thispagestyle{empty}
\pagestyle{empty}

\begin{abstract}
Autonomous systems rely on sensors to estimate the environment around them. However, cameras, LiDARs, and RADARs have their own limitations. In nighttime or degraded environments such as fog, mist, or dust, thermal cameras can provide valuable information regarding the presence of objects of interest due to their heat signature. They make it easy to identify humans and vehicles that are usually at higher temperatures compared to their surroundings. In this paper, we focus on the adaptation of thermal cameras for robotics and automation, where the biggest hurdle is the lack of data. Several multi-modal datasets are available for driving robotics research in tasks such as scene segmentation, object detection, and depth estimation, which are the cornerstone of autonomous systems. However, they are found to be lacking in thermal imagery. Our paper proposes a solution to augment these datasets with synthetic thermal data to enable widespread and rapid adaptation of thermal cameras. We explore the use of conditional diffusion models to convert existing RGB images to thermal images using self-attention to learn the thermal properties of real-world objects. 
\end{abstract}


\begin{figure*}[ht]
  \centering
  \begin{tabular}{|c|c|c|c|c}
    \hline
    \includegraphics[width=3cm, height=2cm]{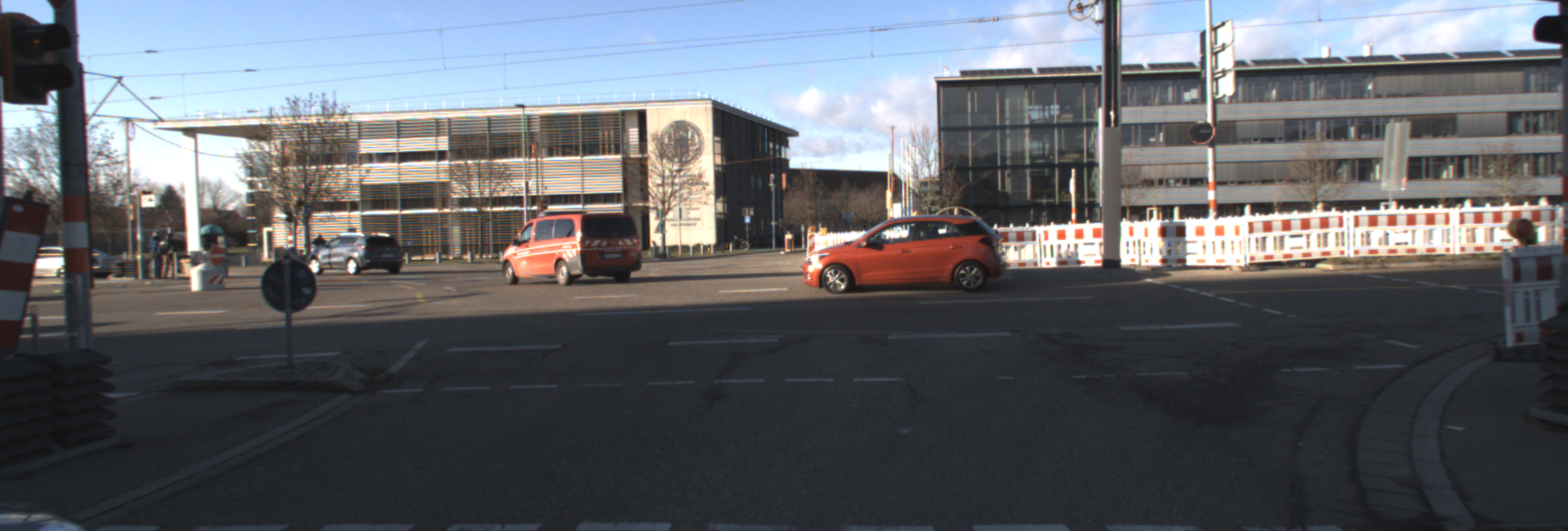} &
    \includegraphics[width=3cm, height=2cm]{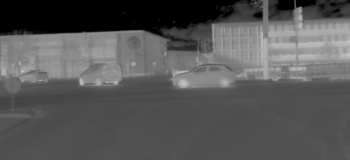} &
    \includegraphics[width=3cm, height=2cm]{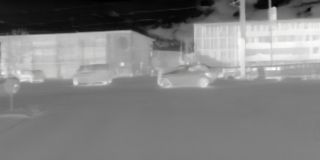} &
    \includegraphics[width=3cm, height=2cm]{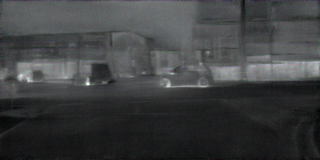} &
    \includegraphics[width=3cm, height=2cm]{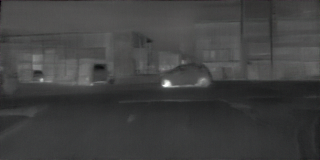}\\
    \hline
    \includegraphics[width=3cm, height=2cm]{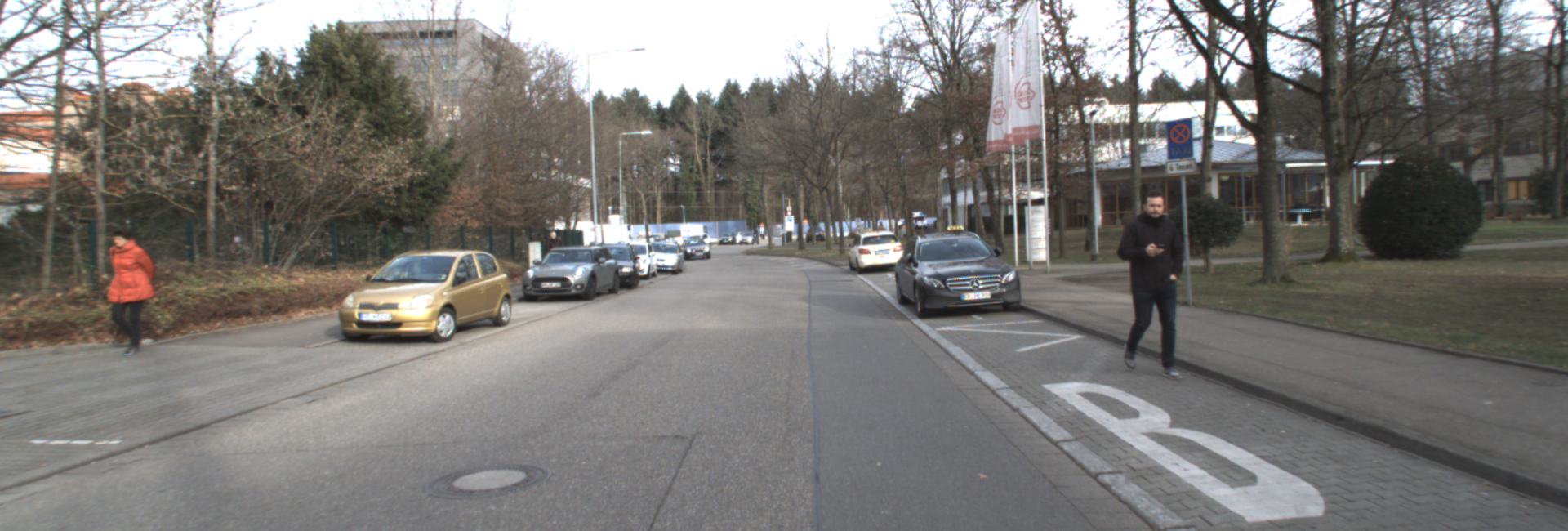} &
    \includegraphics[width=3cm, height=2cm]{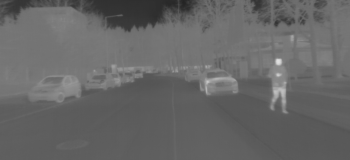} &
    \includegraphics[width=3cm, height=2cm]{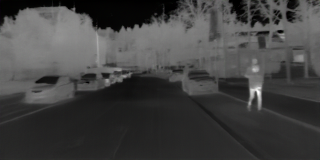} &
    \includegraphics[width=3cm, height=2cm]{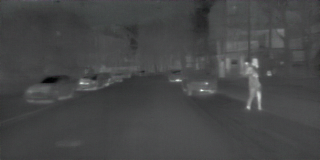} &
    \includegraphics[width=3cm, height=2cm]{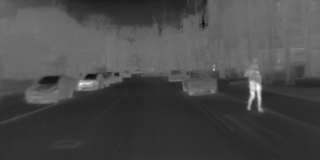} \\
    \hline
    \includegraphics[width=3cm, height=2cm]{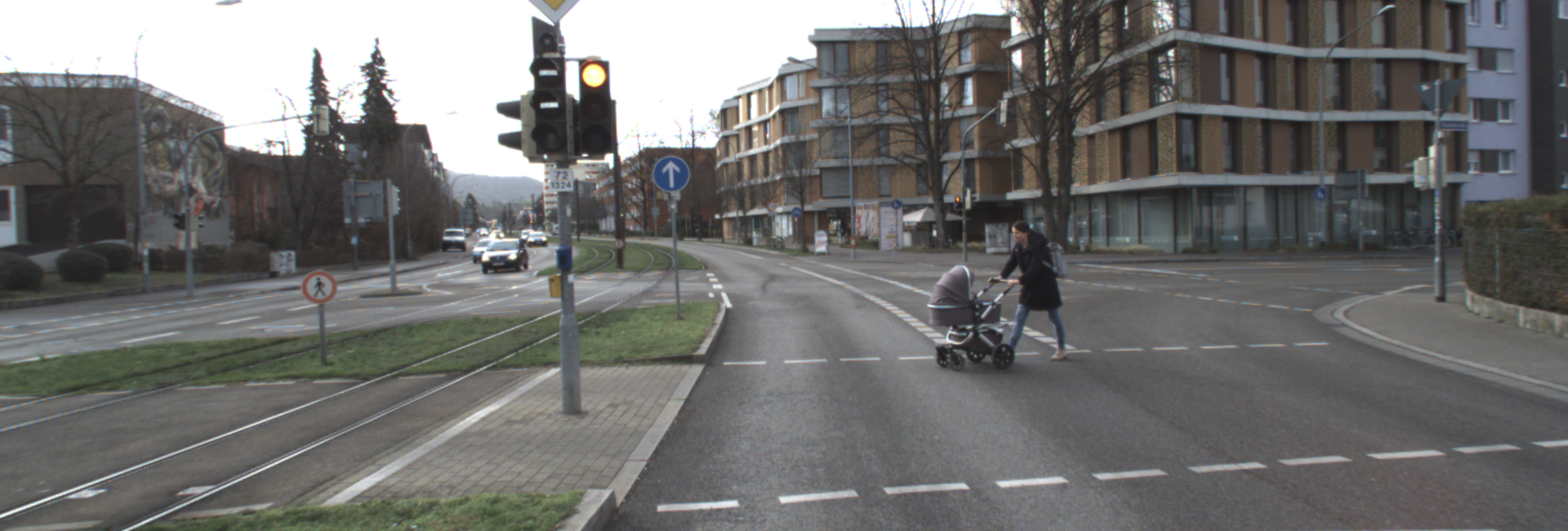} &
    \includegraphics[width=3cm, height=2cm]{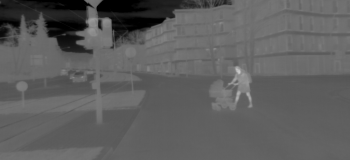} &
    \includegraphics[width=3cm, height=2cm]{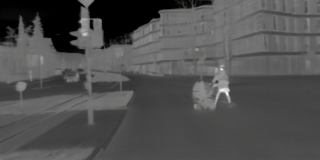} &
    \includegraphics[width=3cm, height=2cm]{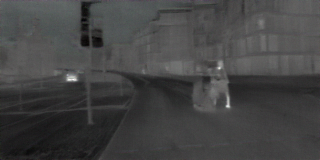} &
    \includegraphics[width=3cm, height=2cm]{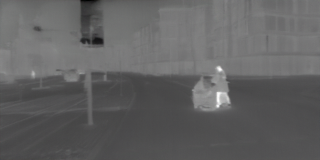} \\
    \hline
    \includegraphics[width=3cm, height=2cm]{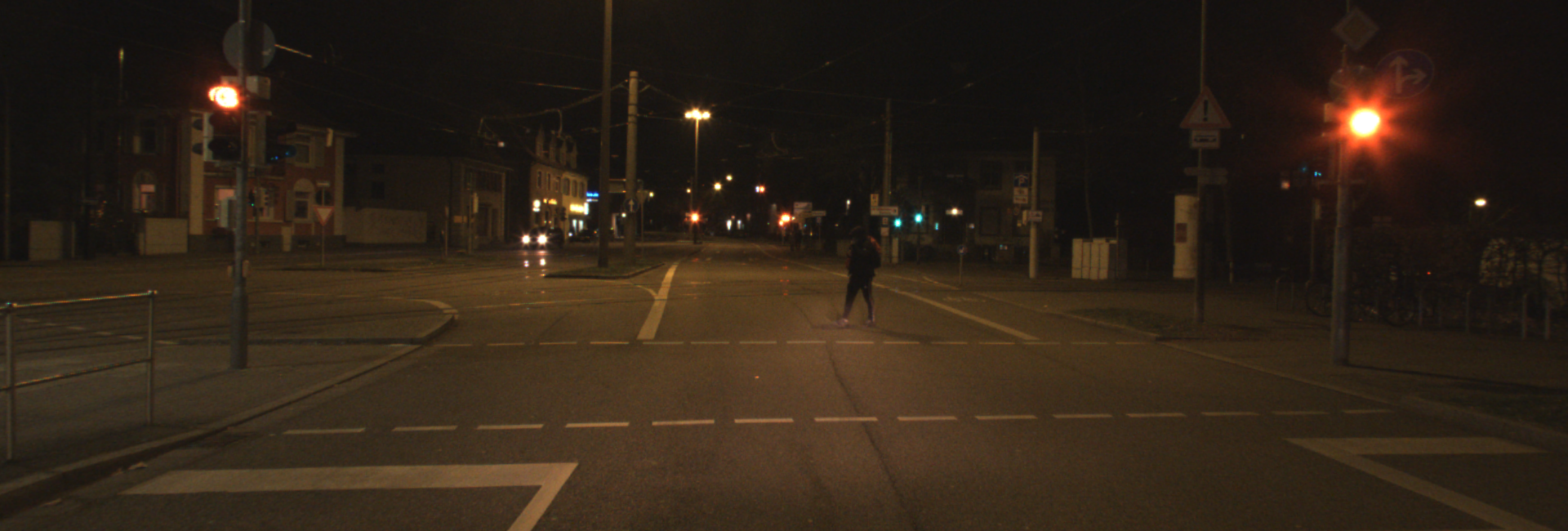} &
    \includegraphics[width=3cm, height=2cm]{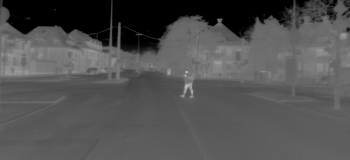} &
    \includegraphics[width=3cm, height=2cm]{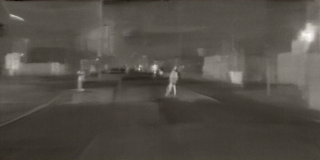} &
    \includegraphics[width=3cm, height=2cm]{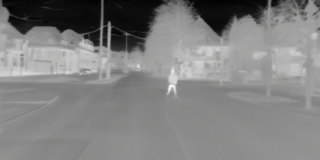} &
    \includegraphics[width=3cm, height=2cm]{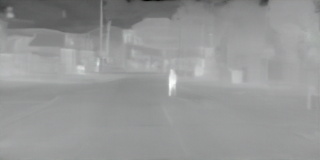} \\
    \hline
    \includegraphics[width=3cm, height=2cm]{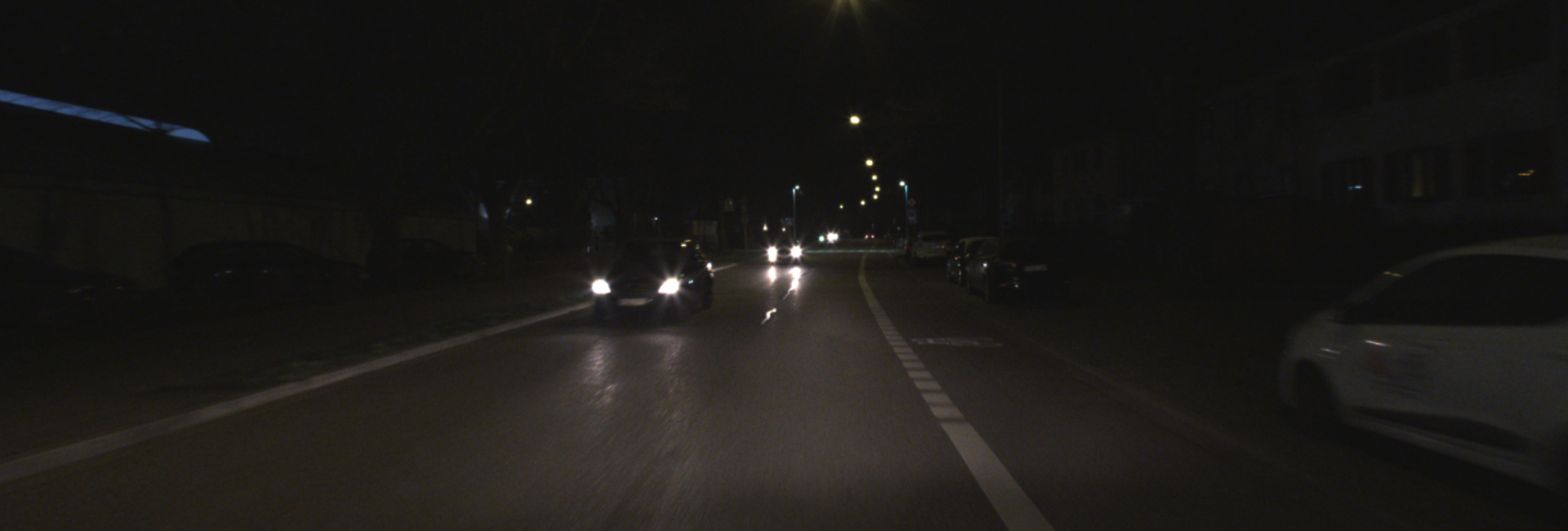} &
    \includegraphics[width=3cm, height=2cm]{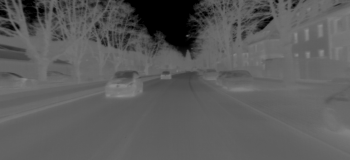} &
    \includegraphics[width=3cm, height=2cm]{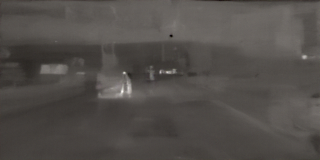} &
    \includegraphics[width=3cm, height=2cm]{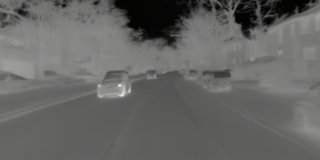} &
    \includegraphics[width=3cm, height=2cm]{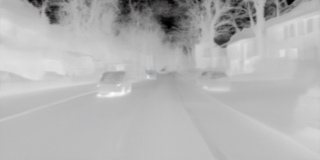} \\
    \hline
    \includegraphics[width=3cm, height=2cm]{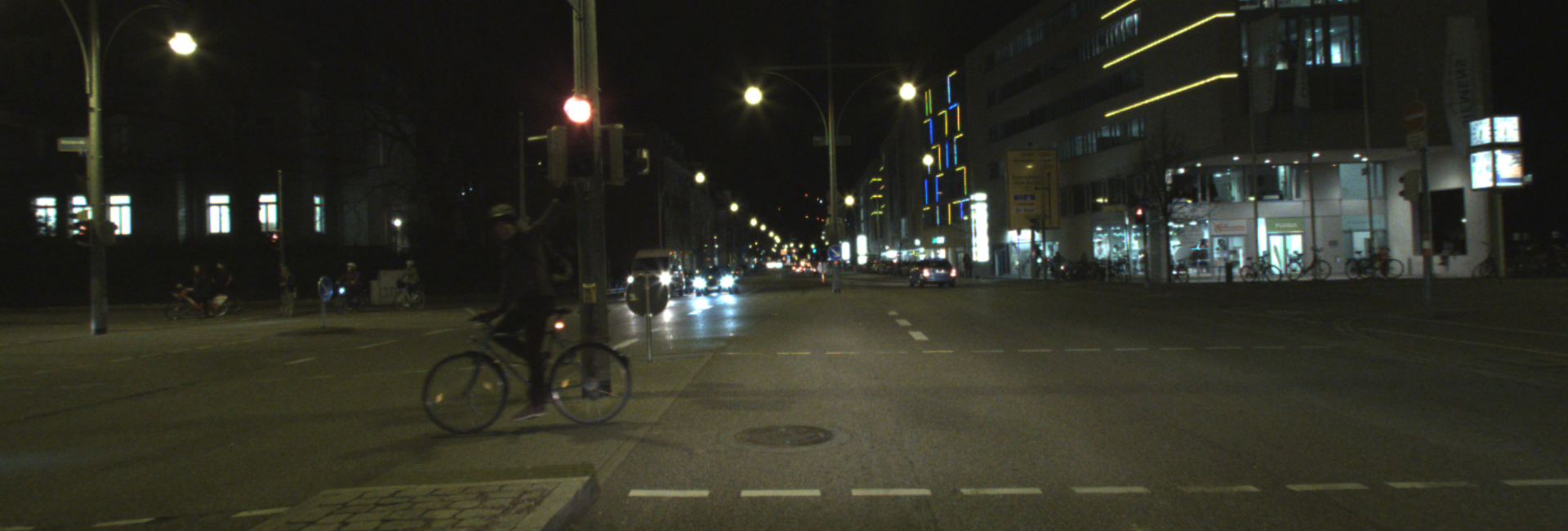} &
    \includegraphics[width=3cm, height=2cm]{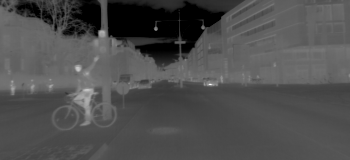} &
    \includegraphics[width=3cm, height=2cm]{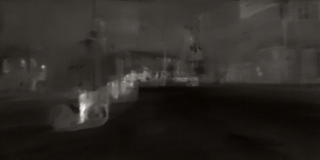} &
    \includegraphics[width=3cm, height=2cm]{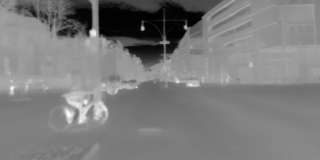} &
    \includegraphics[width=3cm, height=2cm]{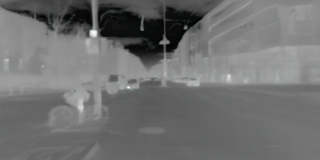} \\
    \hline
    
  \end{tabular}
  \caption{Row 1-3: Freiburg Thermal Daytime Images, Row 4-6: Freiburg Thermal Nighttime Images, Column 1: RGB Images, Column 2: GT Thermal Images, Column 3: Thermal Images for model trained only on daytime data, Column 4:Thermal Images for model trained only on nighttime data, Column 5: Thermal Images for model trained on combined daytime and nighttime data.}
  \label{fig:Freiburg}
\end{figure*}

\section{INTRODUCTION}

Image-to-Image translation in Computer Vision transfers key information from a source domain to a target domain and has widespread application in the domain of Image Processing. Deep learning models, including CNNs, GANs, and diffusion models, have been widely explored for their applications, including image enhancement, super-resolution, data enhancement, colorization, and style transfer. In this paper, we explore the use of Image-to-Image translation to convert RGB data to the thermal domain. 

Thermal images capture the heat patterns emitted by objects in the environment in the form of infrared radiation, where the intensity of the pixels indicates the temperature levels. The ability of thermal images to capture these heat signatures, which are invisible to the naked eye, makes them an incredible tool for various applications such as security and surveillance, firefighting, driving in foggy and smoky conditions, medical examinations, and search and rescue at night. However, obtaining thermal data can be challenging due to the high cost and complexity of the sensors, environmental factors, calibration intricacies, and technical knowledge required.  For robotics applications, most of the datasets today comprise RGB, LIDAR, RADAR, GPS, and IMU data, leaving the need for complementary thermal data. The ability to synthetically generate thermal data for these existing datasets could be a catalyst for the efficient integration of thermal cameras into applications, including Stereo Depth Estimation, Visual Odometry, and SLAM. The availability of high-quality synthetic thermal data corresponding to existing datasets in various domains would enable the supervised training of models for various downstream tasks such as semantic segmentation, object detection, and keypoint detection. This can then enable direct use of thermal cameras for various use cases, such as autonomous driving in urban and off-road conditions, medical imaging, and security. 

Although CNNs have also been explored for image-to-image translations \cite{c2}, conditional GANs have shown great potential in domain conversion \cite{c6} and style transfer \cite{c28, c7}. Lately, however, Denoising Diffusion Probabilistic Models (DDPM) \cite{c20, c1} have emerged as the leaders in high-quality image generation. In this paper, we utilize the principles of conditional diffusion combined with DDPM \cite{c18} to generate Thermal Images from RGB Images. The main improvement over prior works in this domain is the ability of the network to represent salient objects according to their temperature and differentiate between image background and key objects of interest such as pedestrians or vehicles. We build on the guided diffusion proposed in \cite{c1} by Dhariwal et al. and replace their classifier-guided diffusion with the conditional diffusion proposed in \cite{c18}. 

The major contributions of this paper are:
\begin{itemize}
    \item This paper explores the application of Conditional DDPM \cite{c18} for Paired RGB-to-Thermal image translation with a primary focus on autonomous driving and other robotics applications in different environments supported by qualitative and quantitative analysis.
    \item We expand the original attention level proposed in the underlying guided diffusion model \cite{c1} to higher image resolutions, which led to a significant improvement in the identification and correlation of high temperature objects in scenes and also led to an increase in high-frequency information in the form of image details.
    \item We identify the domain gap between daytime and nighttime thermal images with regard to training data and perform an ablation study to maximize the performance of the model. 
    \item We demonstrate the efficacy of our model on multiple datasets including Freiburg \cite{c26}, Caltech Aerial RGB-Thermal Dataset \cite{c14}, KAIST Multispectral Pedestrian \cite{c32}, and FLIR Thermal datasets \cite{flir_data} that contain urban daytime, urban nighttime, and terrestrial data (Caltech). We evaluate the adaptability of our model and compare the similarity of the synthetically generated and GT thermal images. 
    
\end{itemize}

\section{RELATED WORKS}
The task of Image to Image (I2I) translation involves learning a mapping between a source and target domain. It gained momentum with the advent of deep generative models, which aim to model the underlying image data distribution and generate new samples from it. Variational Autoencoders (VAEs) \cite{c4} approach this by maximizing a variational lower bound on the data likelihood through latent-variable modeling. In contrast, Generative Adversarial Networks(GANs) \cite{c5} formulate generation as a two-player game between a generator, which synthesizes images, and a discriminator, which distinguishes real from fake. GANs have become particularly prominent in I2I translation due to their ability to produce sharp, high-fidelity images under paired and unpaired training setups.

\textbf{GAN-based Image Translation.} Conditional GANs guide synthesis with additional information, enabling applications such as inpainting \cite{c27}, style transfer \cite{c28} and superresolution \cite{c29}. Pix2pix \cite{c6} introduced a paired I2I framework using U-Net as the generator and PatchGAN as the Discriminator. CycleGAN \cite{c7} removed the requirement for paired supervision by introducing cycle-consistency loss. Later models like BiCycleGAN \cite{c8}, UNIT \cite{c9}, and MUNIT \cite{c10} explored different latent representations to improve diversity and fidelity.


\textbf{Visual-to-Thermal Translation with GANs.} While GANs, have shown success in various I2I tasks, visual (RGB) to thermal translation presents unique challenges due to its modality gap and limited datasets. ThermalGAN \cite{c11} applied BicycleGAN \cite{bicyclegan} to RGB-Thermal conversion for person re-identification, using two generator networks to synthesize thermal image segmentation and temperature contrast. InfraGAN \cite{c12} added SSIM-based loss for better pixel-level alignment. Other efforts include object-level generators \cite{c22}, CycleGAN-based drone data translation \cite{c7}, and joint optmiziation of image translation and template matching \cite{c35}. However, Wadsworth et al. \cite{c21} showed that even Pix2Pix models trained on outdoor datasets failed to reconstruct essential thermal details such as pedestrians and vehicles. These limitations stem partly from the complexity of modeling physical temperature distributions and partly from challenges inherent to GAN training, such as mode collapse and sensitivity to hyperparameter tuning.

\textbf{Diffusion-Based Image Translation.}
Diffusion models, particularly denoising diffusion probabilistic models (DDPMs), have emerged as strong alternatives to GANs by enabling stable training and improved diversity. Unlike GANs, DDPMs learn to generate data by iteratively denoising from Gaussian noise. UNIT-DDPM \cite{c17} extends this to unpaired I2I translation by using joint distribution modeling. Although it includes RGB-thermal examples, the results lack details and heat signature of objects. Other works on guided diffusion techniques \cite{c1, c18} enable conditional generation but often trade off image diversity. 
Palette \cite{c20} highlights the importance of self-attention in diffusion-based I2I tasks, showing competitive results in inpainting and colorization. We extends upon the idea to RGB-Thermal translation.


\textbf{Visual-to-Thermal Translation with Diffusion Models.} PID \cite{c16} incorporates physical modeling by decomposing thermal images into temperature and emissivity components, and combines with latent diffusion to generate physically plausible thermal imagery. While our approach is not explicitly physics-based, our conditional diffusion model leverages self-attention to learn object-specific thermal characteristics, implicitly correlating visual appearance with temperature, such as warmer human bodies or heated tires from friction.

T2V-DDPM \cite{c15} performs thermal-to-visible translation for facial imagery using conditional DDPM. Building upon similar principles, our work focuses on the reverse tasks, visible-to-thermal translation, in dynamic outdoor environments for autonomous navigation and extend the self-attention resolutions described in \cite{c1, c20}.

\begin{figure*}[ht]
  \centering
  \begin{tabular}{|c|c|c|c|c}
    \hline
    \includegraphics[width=3cm, height=2cm]{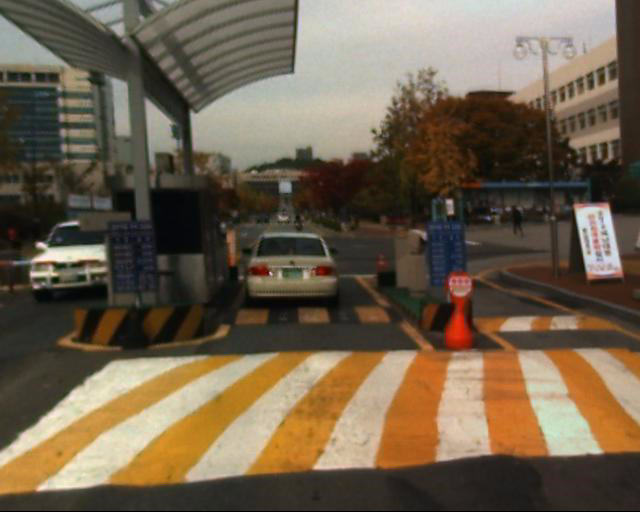} &
    \includegraphics[width=3cm, height=2cm]{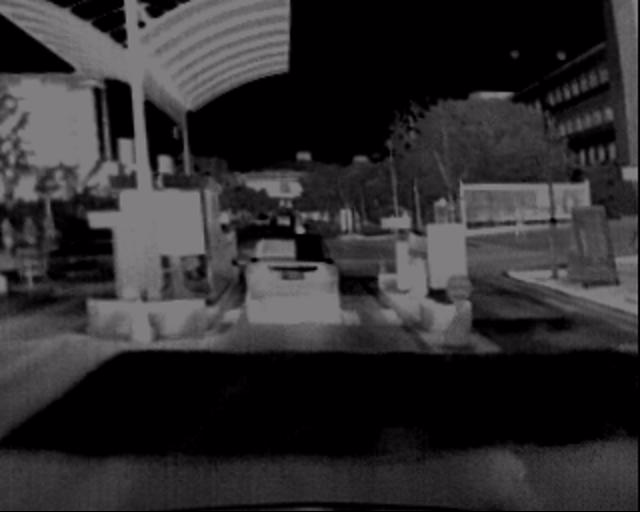} &
    \includegraphics[width=3cm, height=2cm]{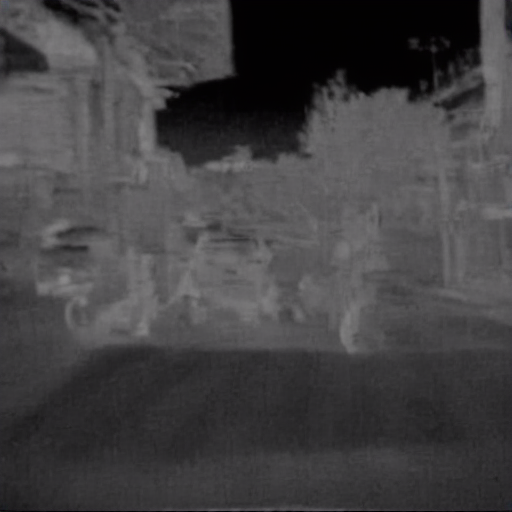} &
    \includegraphics[width=3cm, height=2cm]{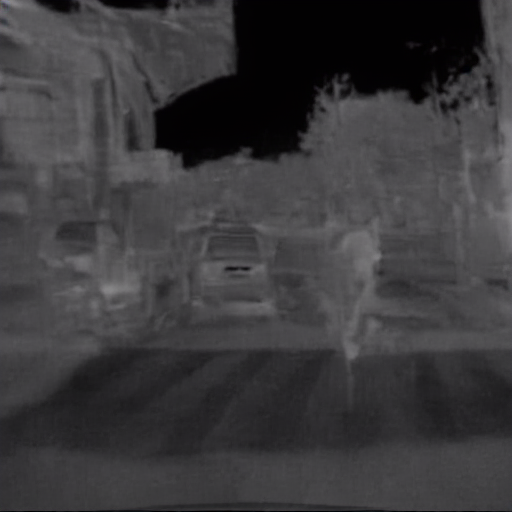} &
    \includegraphics[width=3cm, height=2cm]{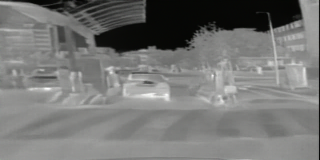} \\
    \hline
    \includegraphics[width=3cm, height=2cm]{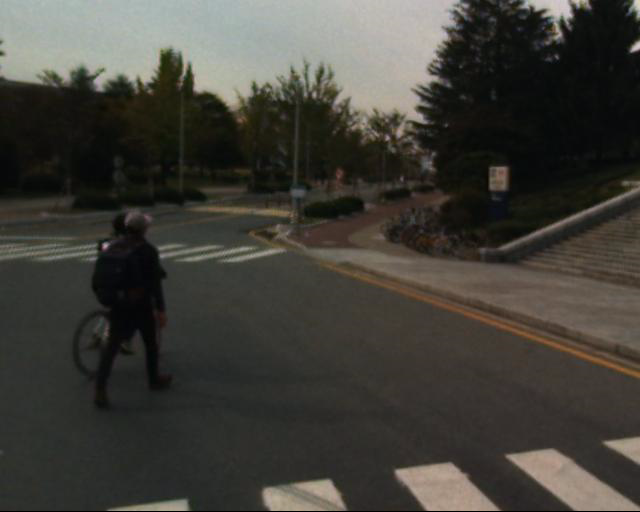} &
    \includegraphics[width=3cm, height=2cm]{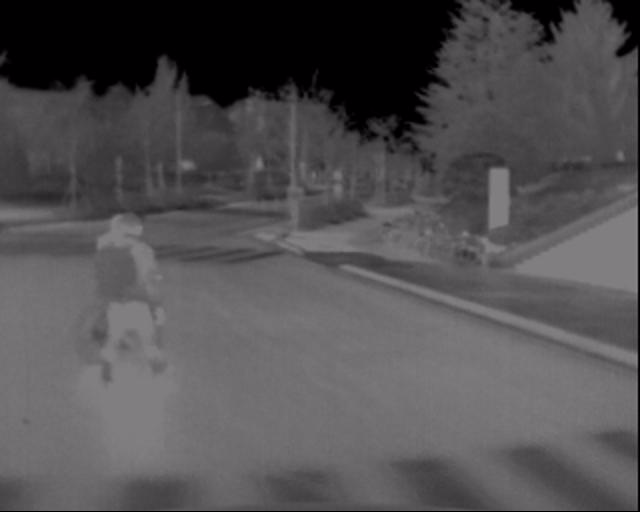} &
    \includegraphics[width=3cm, height=2cm]{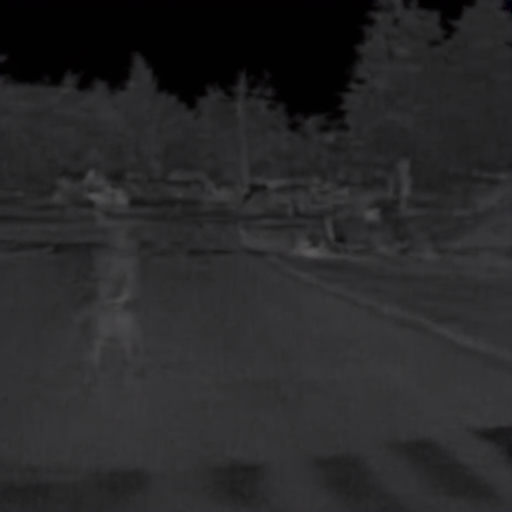} &
    \includegraphics[width=3cm, height=2cm]{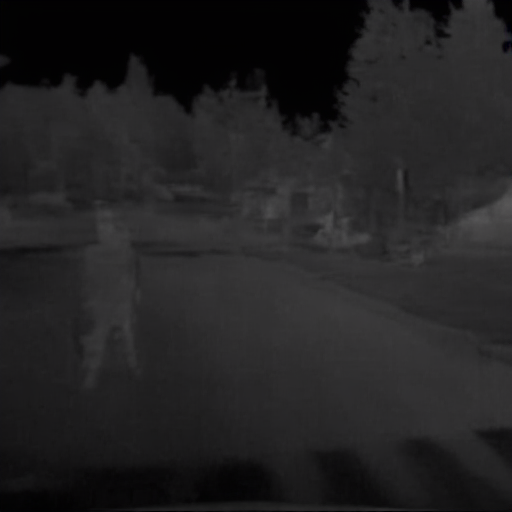} &
    \includegraphics[width=3cm, height=2cm]{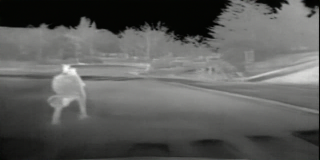} \\
    \hline
    \includegraphics[width=3cm, height=2cm]{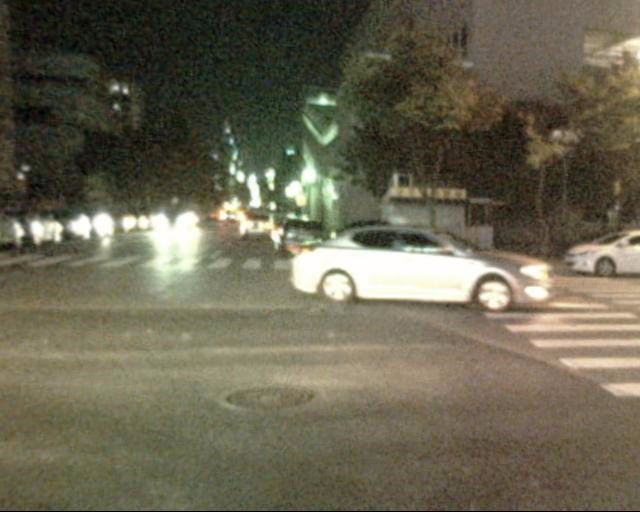} &
    \includegraphics[width=3cm, height=2cm]{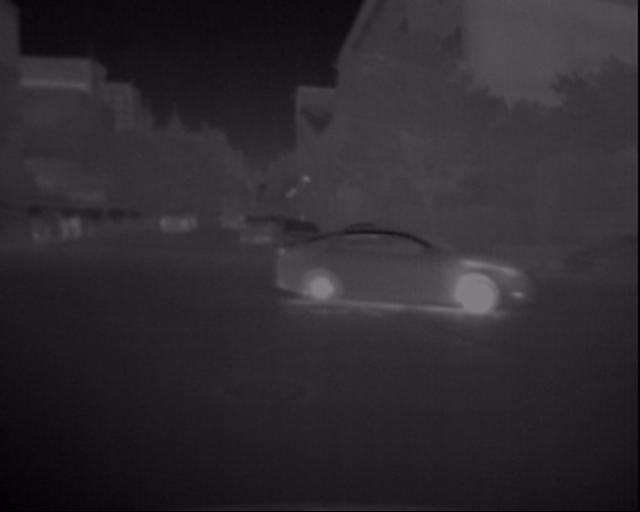} &
    \includegraphics[width=3cm, height=2cm]{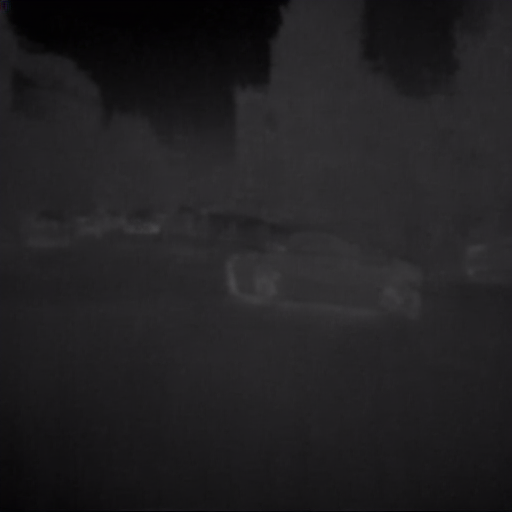} &
    \includegraphics[width=3cm, height=2cm]{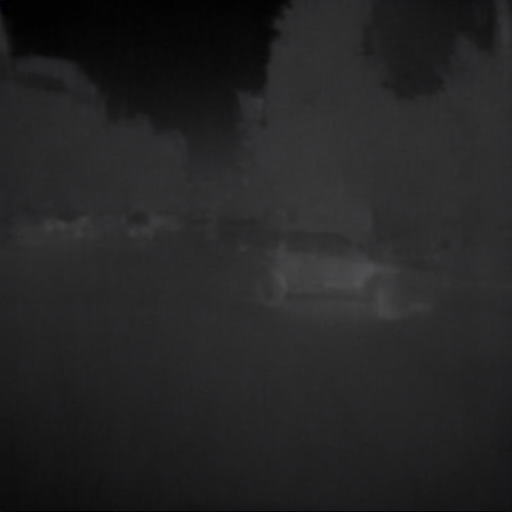} &
    \includegraphics[width=3cm, height=2cm]{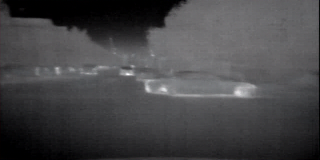} \\
    \hline
    \includegraphics[width=3cm, height=2cm]{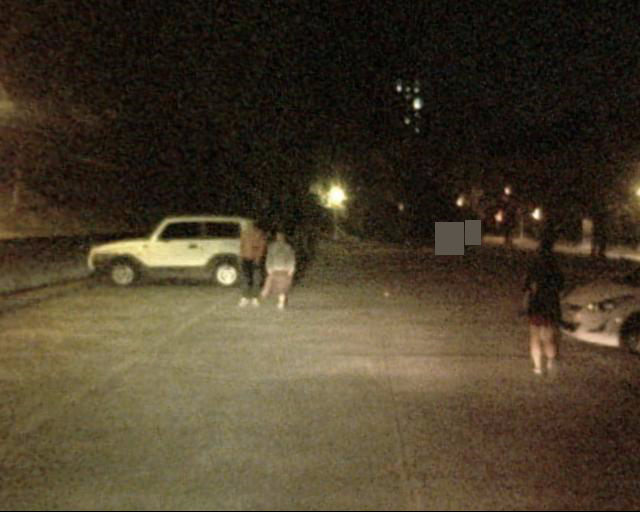} &
    \includegraphics[width=3cm, height=2cm]{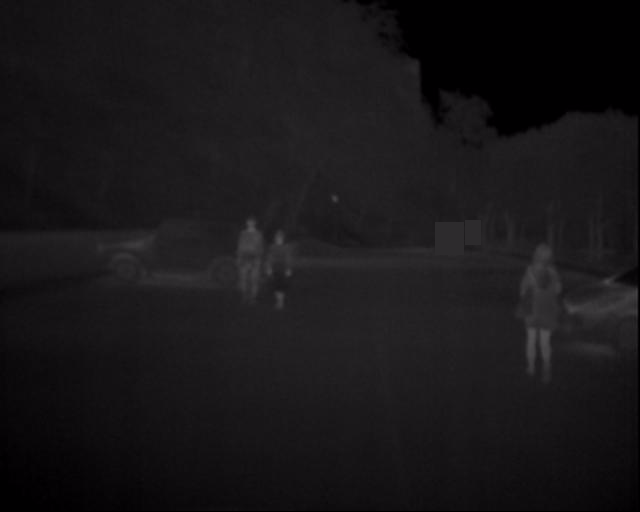} &
    \includegraphics[width=3cm, height=2cm]{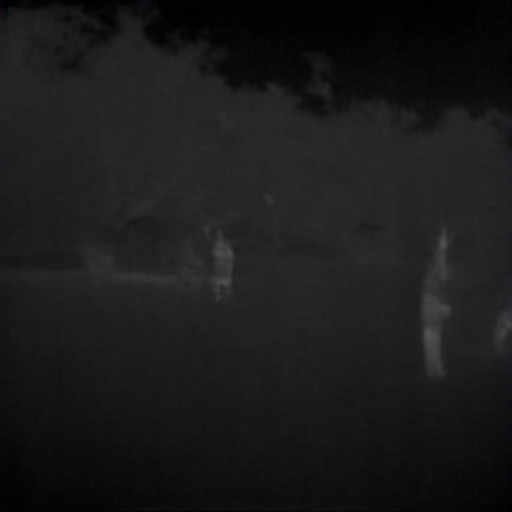} &
    \includegraphics[width=3cm, height=2cm]{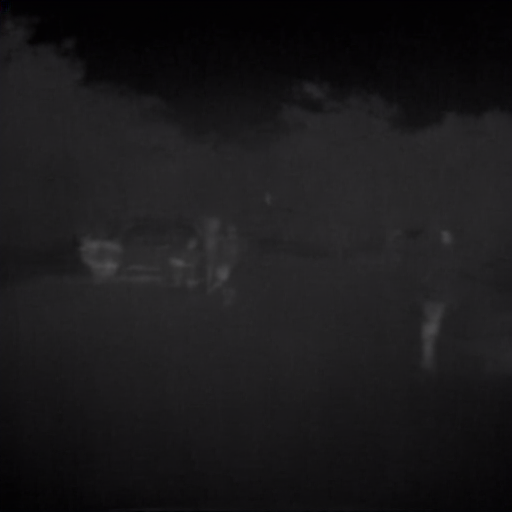} &
    \includegraphics[width=3cm, height=2cm]{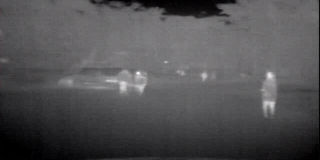} \\
    \hline

  \end{tabular}
  \caption{Row 1-3: KAIST Daytime Images, Row 4-6: KAIST Nighttime Images, Row 7-9; Column 1: RGB Images, Column 2: GT Thermal Images, Column 3: Thermal Images (PID \cite{c16}) , Column 4: Thermal Images (LDM \cite{c25}), Column 5: Ours.}
  \label{fig:KAIST}
\end{figure*}
\section{METHODOLOGY}

\subsection{Denoising Diffusion Probabilistic Model (DDPM)}

In Denoising Diffusion Probabilistic Models \cite{c24}, the image generation process is divided into a forward and a reverse process. In the forward process, we take the ground truth image, \(y_0 \) and incrementally add Gaussian noise to it in each step to obtain the next noisier image. The image at the t-th instance, \(y_t \) instance given the image at time t-1, \(y_{t-1} \) can be represented by a Markovian process, q.

\begin{equation}
q(y_{1:T} \mid y_0) = \prod_{t=1}^{T} q(y_t \mid y_{t-1})
\end{equation}
\begin{equation}
q(y_t \mid y_{t-1}) = \mathcal{N}(y_t \mid \sqrt{\alpha_t} y_{t-1}, (1 - \alpha_t)I)
\end{equation}

The reverse process starts with a pure Gaussian noise image, \(y_t\) and in each reverse step, performs an iterative denoising operation to finally obtain a high-quality image, \(y_0\) according to the training distribution. Given a large number of timesteps and a small variance of additive Gaussian noise, the reverse step going from time instant \(y_t\) to \(y_{t-1}\) can be approximated by a Gaussian as:

\begin{equation}
p(y_{t-1} \mid y_t, y_0) = \mathcal{N}\left(y_{t-1}; \mu, \sigma^2 I\right)
\end{equation}

\begin{equation}
\mu = \frac{ \sqrt{\gamma_{t-1}}(1 - \alpha_t) }{1 - \gamma_t} y_0 + \frac{ \sqrt{\alpha_t} (1 - \gamma_{t-1})}{1 - \gamma_t} y_t
\end{equation}

\begin{equation}
\sigma^2 = \frac{(1 - \gamma_{t-1})(1 - \alpha_t)}{1 - \gamma_t}
\end{equation}

Here, \(\gamma_t\) is the product of noise schedule parameters, \(\alpha\) up to time t: \(\gamma_t = \prod_{i=1}^{t} \alpha_i\)

The parameters of the reverse process are modeled by a neural network given by:

\begin{equation}
p_\theta(x_{t-1} \mid x_t) = \mathcal{N}\left( \mu_\theta(x_t, t), \Sigma_\theta(x_t, t) \right)
\end{equation}

Keeping the noise variance the same during the backward and forward process, only the mean values need to be modeled for each step of the reverse process. For this, typically the U-Net model \cite{c3} is used.

\subsection{Conditional DDPM}
For conditional image generation, we have a source image domain $x$ and a target image domain $y$. We need to learn the target distribution \( p(y\mid x)\) by adapting the DDPM for this translation. According to Saharia et al. \cite{c18}, we can apply x as a constraint to the DDPM to make it learn the conditional distribution according to the equations:

\begin{equation}
p_\theta(y_{0:T} \mid x) = p(y_T) \prod_{t=1}^{T} p_\theta(y_{t-1} \mid y_t, x)
\end{equation}
\begin{equation}
p_\theta(y_{t-1} \mid y_t, x) = \mathcal{N}\left(y_{t-1} \mid \mu_\theta(x, y_t, \gamma_t), \sigma_t^2 I \right)
\end{equation}

The reverse diffusion process goes through an iterative refinement process, where the source image is our constraint, x, and the target image is \(y_0\). It starts with a pure noisy image as discussed in DDPM above; however, now each step in the recovery process of obtaining the noiseless image \(y_0\) is conditioned on the noisy \(y_t\) image from the previous step and the source image, x. The noise to be subtracted from the y image is calculated using the neural network model,  where x and \(y_t\) are the inputs and the output is the estimated noise. In our case , the target image \(y_0\) is the desired thermal image and the source image, x is the RGB image. 
We apply this conditioning to the base model proposed by Dhariwal et al in \cite{c1} , replacing the classifier guided diffusion which is more appropriate for training images belonging to a specific object class. Here, we want to use the source image to guide the diffusion to generate thermal images corresponding to the input RGB images.

\begin{figure*}[ht]
  \centering
  \begin{tabular}{|c|c|c|c}
    \hline
    \includegraphics[width=3cm, height=2cm]{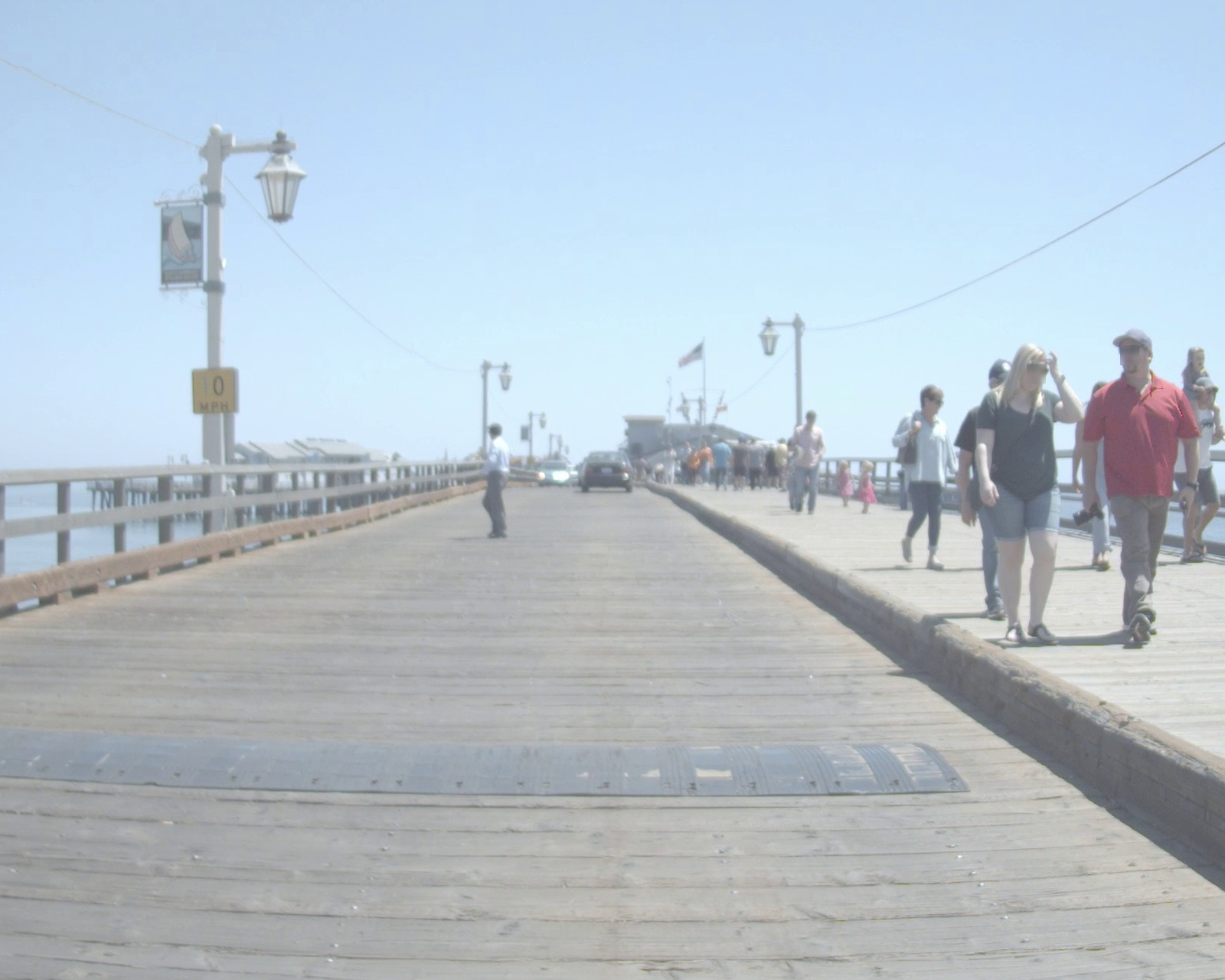} &
    \includegraphics[width=3cm, height=2cm]{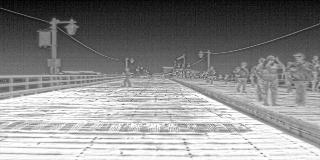} &
    \includegraphics[width=3cm, height=2cm]{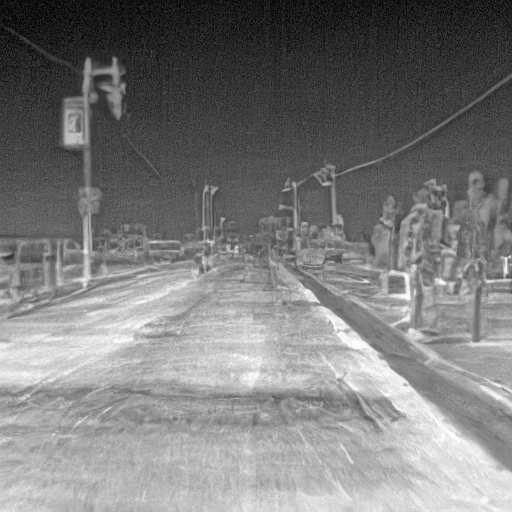} &
    \includegraphics[width=3cm, height=2cm]{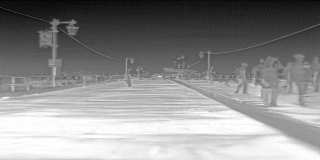} \\
    \hline
    \includegraphics[width=3cm, height=2cm]{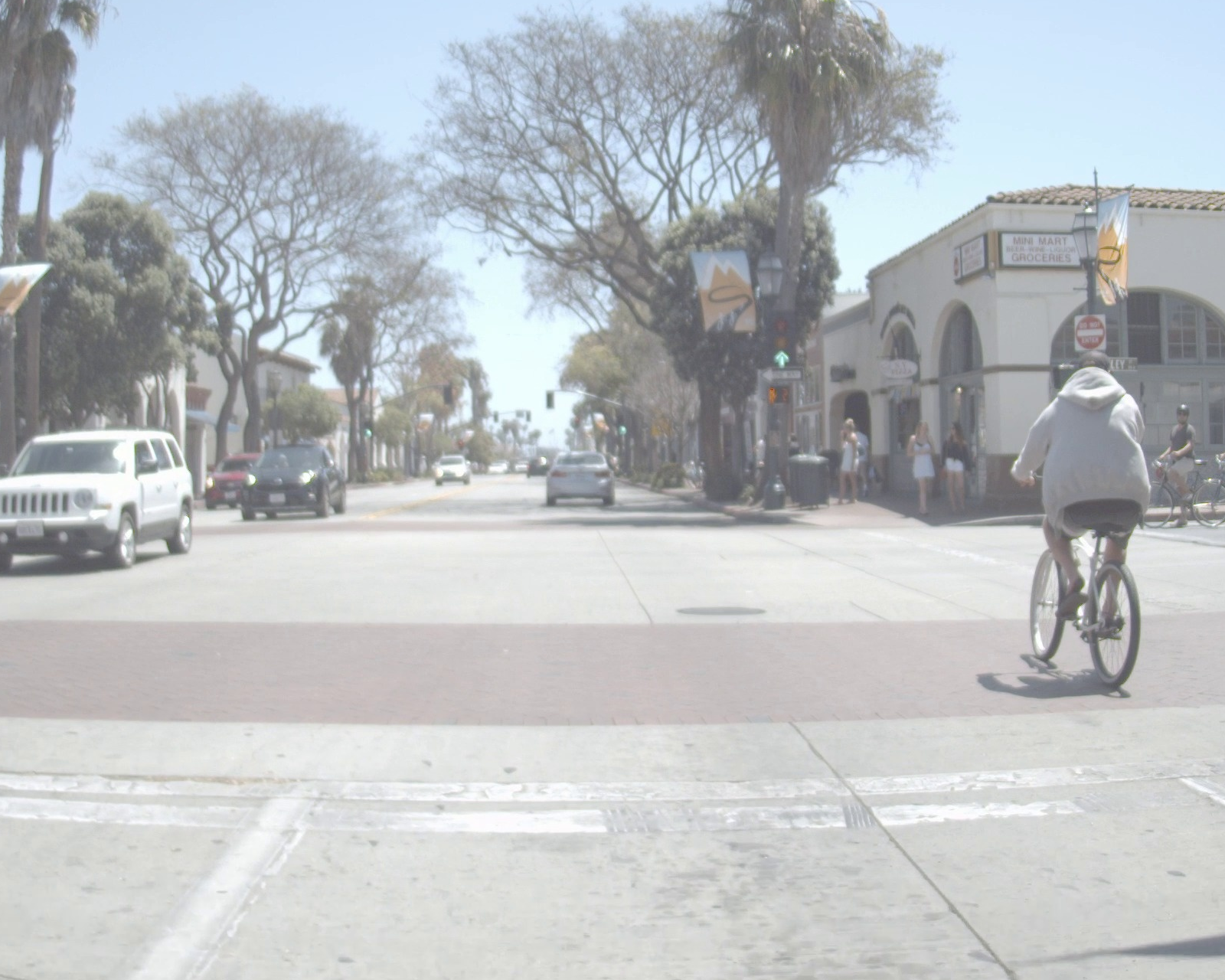} &
    \includegraphics[width=3cm, height=2cm]{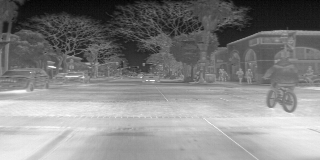} &
    \includegraphics[width=3cm, height=2cm]{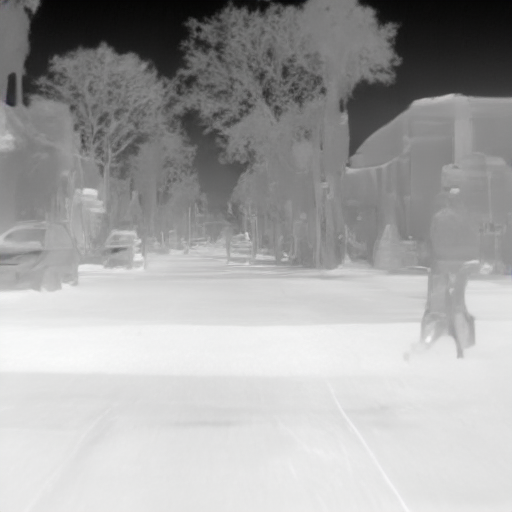} &
    \includegraphics[width=3cm, height=2cm]{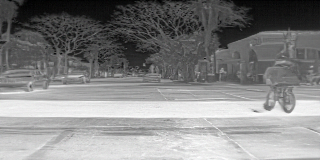} \\
    \hline
    \includegraphics[width=3cm, height=2cm]{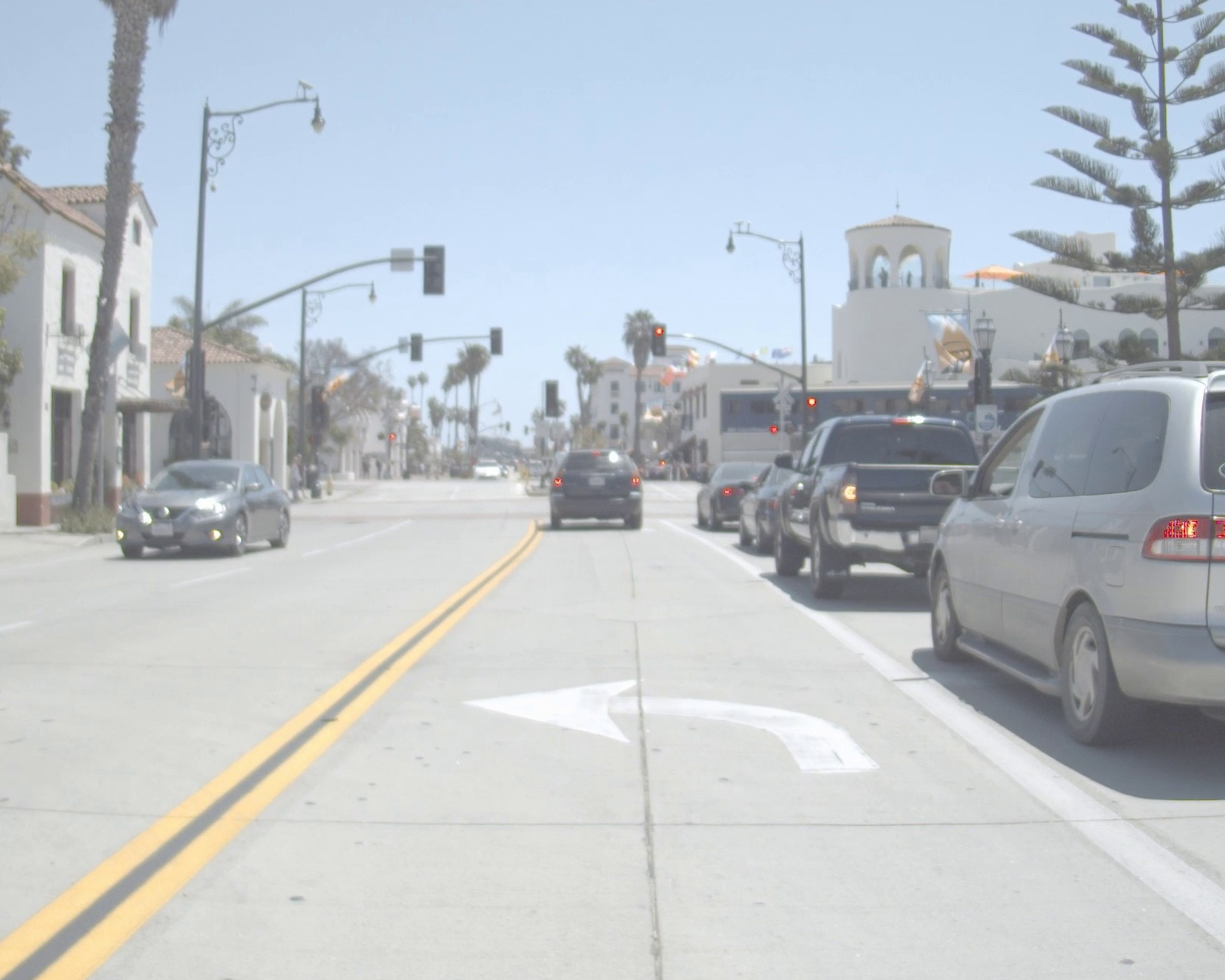} &
    \includegraphics[width=3cm, height=2cm]{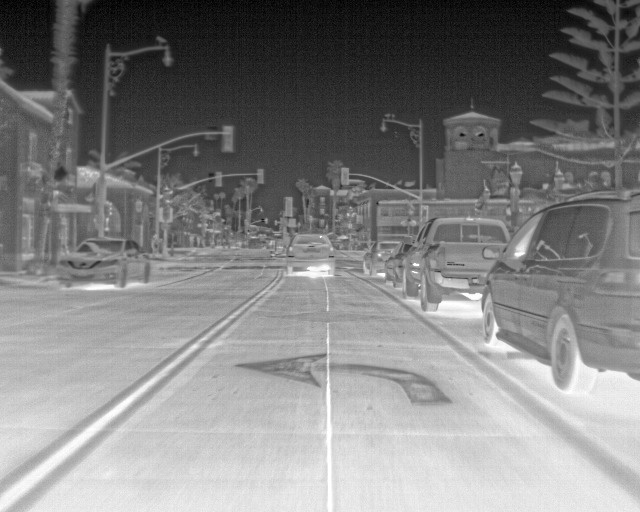} &
    \includegraphics[width=3cm, height=2cm]{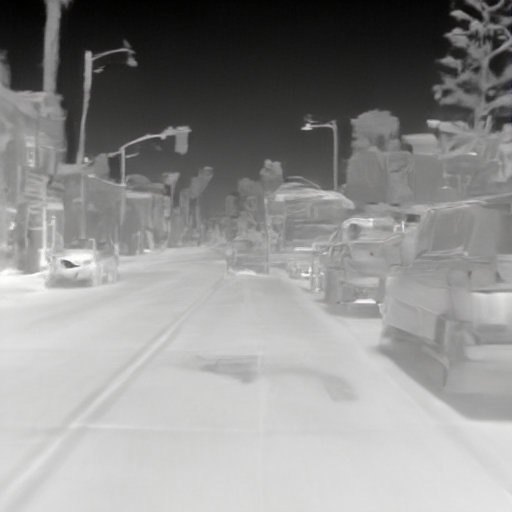} &
    \includegraphics[width=3cm, height=2cm]{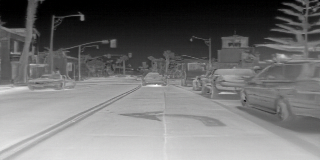} \\
    \hline
    
  \end{tabular}
  \caption{FLIR Dataset Images; Column 1: RGB Images, Column 2: GT Thermal Images, Column 3: Thermal Images (PID) , Column 4: Ours.}
  \label{fig:FLIR}
\end{figure*}

\subsection{The role of self-attention}

DDPM \cite{c24} introduces the use of U-Nets in the neural network backbone for Diffusion models with self-attention at the 16 x 16 resolution for  a 128 x 128 input resolution. Self -attention helps the model understand the correlation between objects in an image. Multi-headed attention helps the model to extend this capability to various aspects of the image. In Stable Diffusion \cite{c25}, the authors emphasize that self-attention helps to preserve the geometric and shape details of the source image while converting to the target image. In several recent works, the importance of self-attention to capture the salient features required for the generation process is reiterated. Guided Diffusion \cite{c1} model increases the number of attention heads and uses attention at 32 x 32, 16 x 16 and 8 x 8 for a 128 x 128 input resolution.
In our implementation, we extend the attention to higher image resolutions of (height/2 x width/2) to help the model learn the correlations between the objects in the images and their heat signatures since the implicit details of thermal images such as higher tire temperature of moving vehicles and higher temperature of the human bodies would be difficult for the model to learn at lower resolutions. In turn, we are rewarded by improved quality thermal images which justify the thermal qualities of objects in the real world. We are, however, constrained by the increasing compute requirements from further increasing attention levels. Training and inference times are significantly longer and memory usage increases as well.

\subsection{Daytime vs Nighttime}

There is a substantial distribution shift in thermal image imagery between day and night, primarily driven by variations in the thermal radiation emitted by objects under different environmental conditions.As demonstrated in \cite{c33}, a 24-hour study of thermal signatures, daytime thermal images show a wider spread of intensity, resulting in more contrast in processed images, while nighttime images concentrate most of the pixel intensity in a narrow range. In particular, vegetation and soil exhibit nearly identical thermal intensity from 20:00 to 6:00, while during the daytime exhibit 2-3 times greater contrast due to solar heating. Although water maintains relatively stable kinetic temperatures, its apparent thermal intensity reverses completely between day and night, appearing cooler than surrounding terrain in daylight but warmer after sunset due to slower heat dissipation. Additionally, solar reflections and thermal shadows that are common during the day further differentiate the daytime thermal image characteristics from those captured at night. 

These shifts in thermal distribution can degrade a model's generalization performance when trained on a mixture of day and night data. Training separate models for daytime and nighttime data better accommodates the distinct radiometric and semantic properties of each period, leading to more stable and accurate image translation. As shown in Figure \ref{fig:Freiburg} and Table \ref{tab:DayandNight}, models trained and evaluated on data from the same period (day or night) outperform those trained across periods or on the combined dataset in terms of PSNR, SSIM, and FID. One exception arises where the combined model performs worse when evaluated on nighttime data. This may stem from the dominance of higher-contrast daytime samples during training, which biases the model toward sharper intensity transitions and broader dynamic range. As a result, the combined model tends to have a larger variance for contrast (as compared to time-specific model). At nighttime, the thermal image has a low variance due to which the combined model leads to a degraded structural alignment and pixel-wise accuracy. Both of these directly affect SSIM and PSNR. In future work, we plan to ablate the impact of training-time distribution bias by explicitly comparing predicted and ground truth thermal histograms, particularly to quantify contrast compression or overestimation in nighttime scenes.
\vspace{0.5cm}
\begin{table}[h!]
\centering
\begin{tabular}{|c|c|c|c|c|}
\hline
\cline{2-5} & & \multicolumn{3}{|c|}{Training Data}\\

\cline{3-5} Test Data & Metric & Day & Night & Day + Night \\
\hline
\multirow{3}{*}{Day} & PSNR $\uparrow$ & 16.93 & 16.10  & 16.48\\
                                   \cline{2-5}
                         & SSIM $\uparrow$ & 0.72 & 0.61  & 0.69 \\
                                   \cline{2-5}
                         & FID  $\downarrow$ & 137.67 & 231.46  & 187.04 \\

\hline

\multirow{3}{*}{Night} & PSNR $\uparrow$ & 14.32 &  14.73  & 13.86\\
                                   \cline{2-5}
                            & SSIM $\uparrow$ & 0.64 & 0.67  &  0.62\\
                                   \cline{2-5}
                            & FID  $\downarrow$ & 241.16 & 162.44  & 164.89\\

\hline

\hline
\end{tabular}
\caption{Quantitative assessment of Daytime and Nighttime Training scenarios for Freiburg Dataset.}
\label{tab:DayandNight}
\end{table}

\begin{figure*}[ht]
  \centering
  \begin{tabular}{|c|c|c|c|}
    \hline
    \includegraphics[width=3cm, height=2cm]{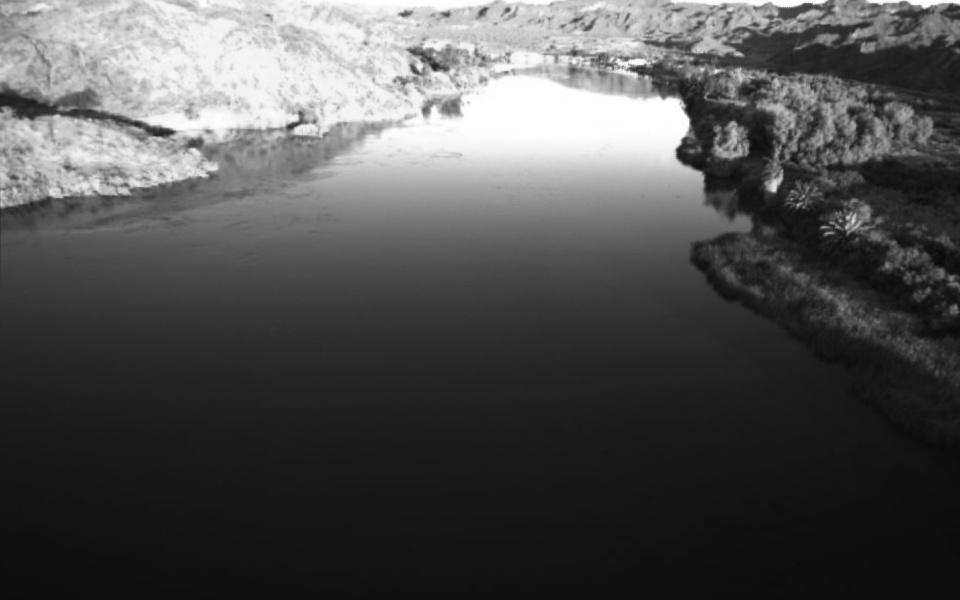} &
    \includegraphics[width=3cm, height=2cm]{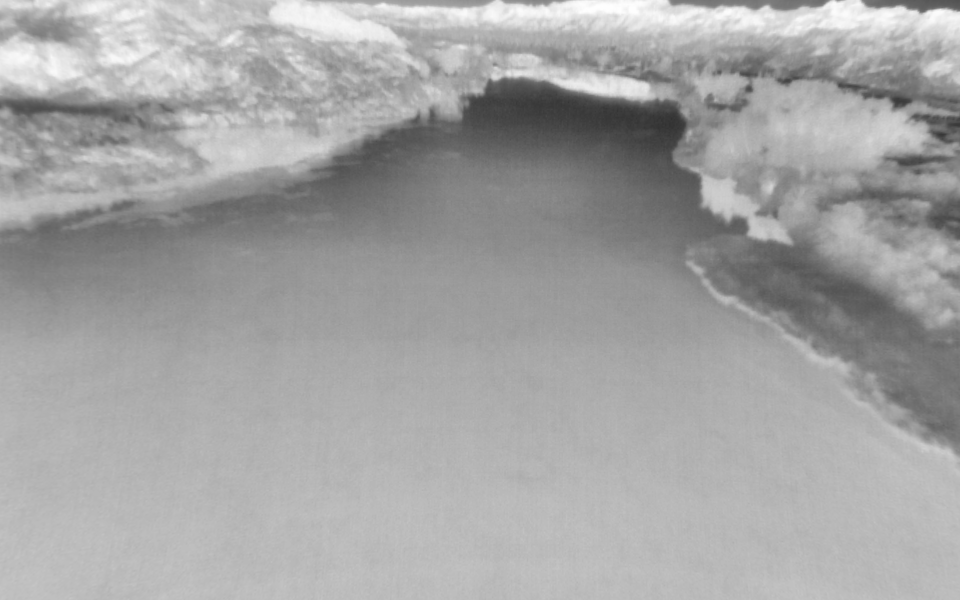} &
    \includegraphics[width=3cm, height=2cm]{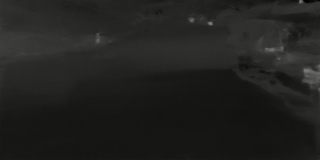} &
    \includegraphics[width=3cm, height=2cm]{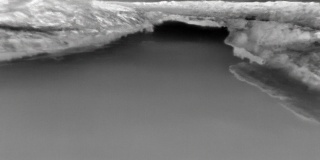} \\
    \hline
    \includegraphics[width=3cm, height=2cm]{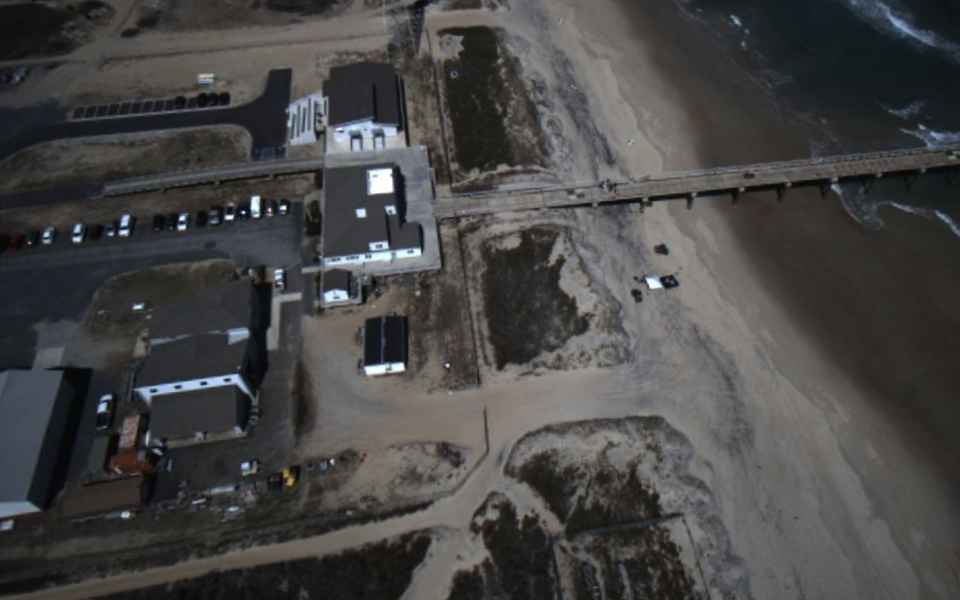} &
    \includegraphics[width=3cm, height=2cm]{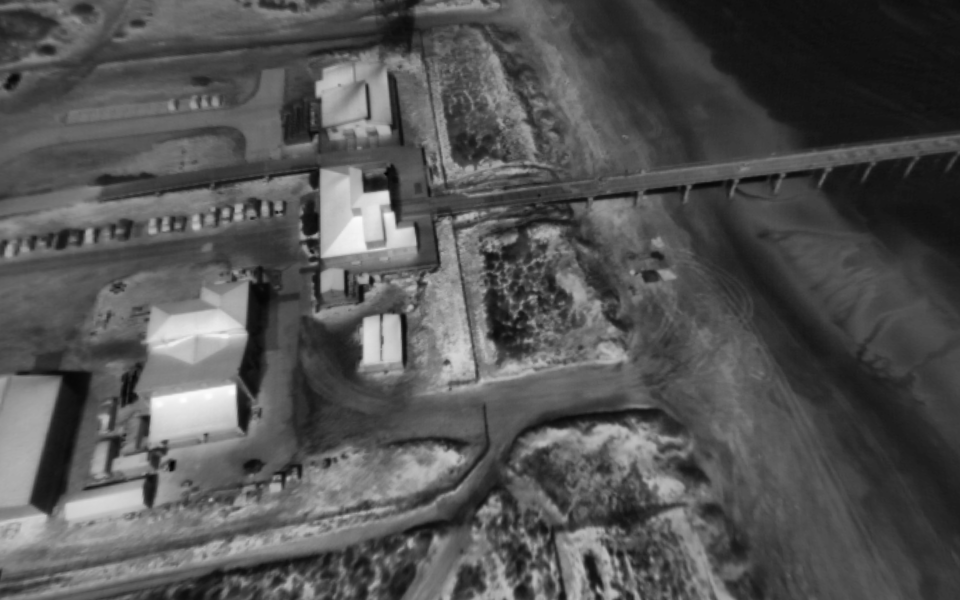} &
    \includegraphics[width=3cm, height=2cm]{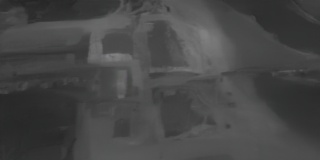} &
    \includegraphics[width=3cm, height=2cm]{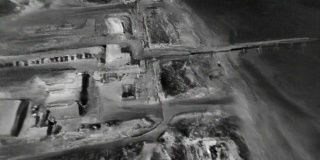} \\
    \hline
    \includegraphics[width=3cm, height=2cm]{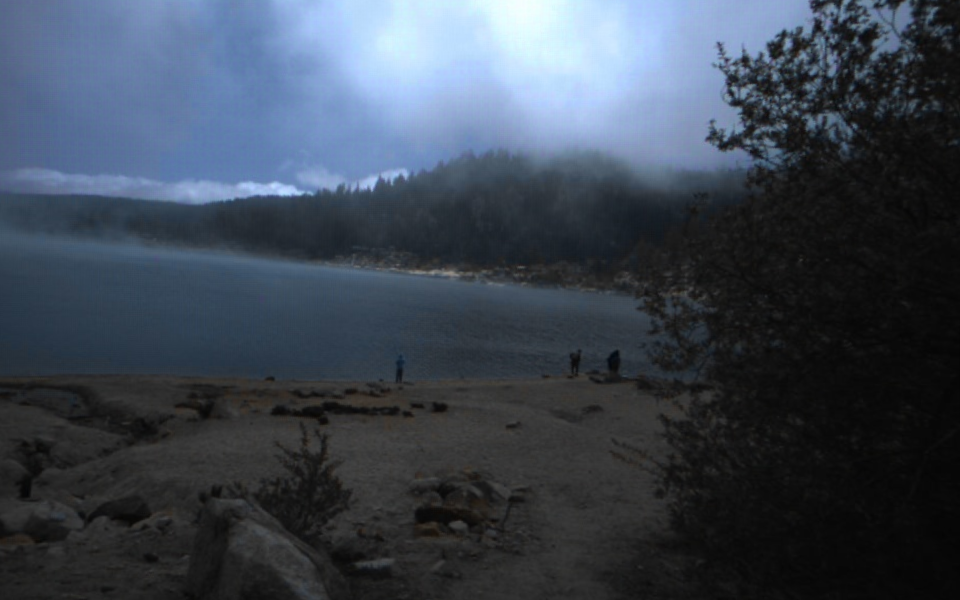} &
    \includegraphics[width=3cm, height=2cm]{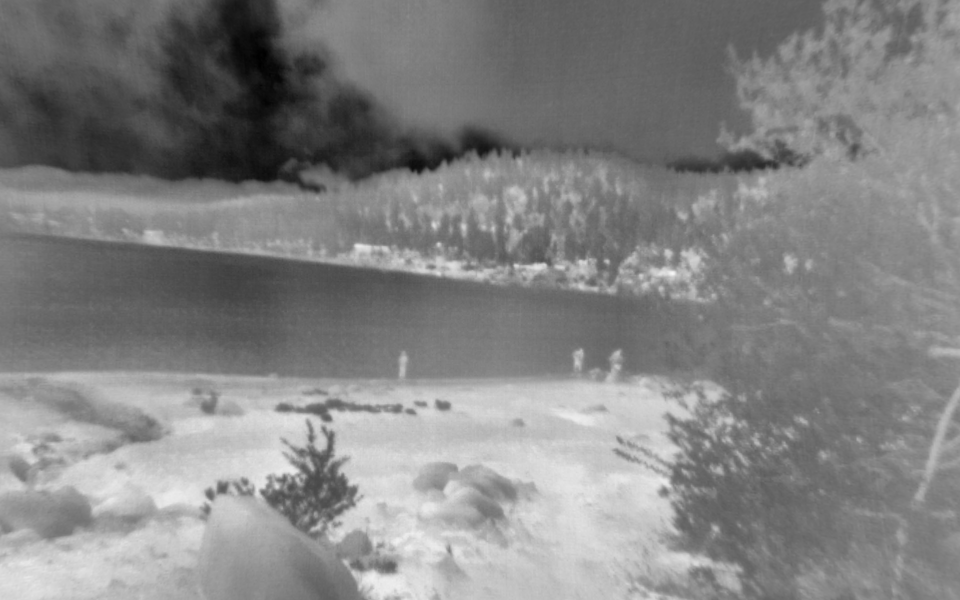} &
    \includegraphics[width=3cm, height=2cm]{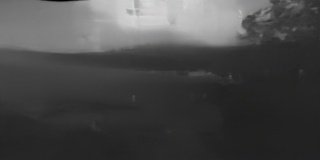} &
    \includegraphics[width=3cm, height=2cm]{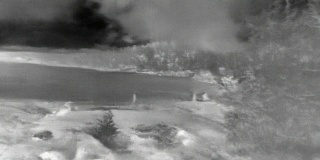} \\
    \hline
    
  \end{tabular}
  \caption{Caltech Aerial Dataset Images; Column 1: RGB Images, Column 2: GT Thermal Images, Column 3: Generated Thermal Images before finetuning (trained on Freiburg dataset), Column 4: Generated Thermal Images after finetuning on Caltech Dataset.The model learns to de-emphasize geometric patterns in the RGB modality in favor of thermal signatures (middle row). As an example, RGB-specific geometric features like waves which are visible in the middle-row RGB (top-right part of the photo) and absent in the thermal modality are accordingly not reproduced in the predicted thermal image.}
  \label{fig:CalTech}
\end{figure*}

\section{EXPERIMENTS}

\subsection{Training on Paired RGB-Thermal Data: Freiburg Thermal Dataset}

\label{sec_exp_freiburg}

We train and test the model on the Freiburg Thermal Dataset \cite{c26}, which contains 20,647 images. The RGB and thermal image pairs in this dataset are time-synchronized and spatially aligned, which facilitates supervised learning of a direct mapping between the two modalities.

After training, the results shown in Figure \ref{fig:Freiburg} demonstrate strong performance. It displays examples from the test set of the Freiburg dataset during day and night time, where the model effectively highlights salient objects within each scene. This is particularly important for applications such as autonomous driving, where detecting pedestrians, vehicles, and other nearby objects is critical.

Our results maintain a high fidelity of thermal characteristics, such as the elevated temperature of vehicle tires due to friction and the warmer body temperatures of pedestrians, both of which are captured accurately by the model. The results are better for models trained and tested on data from the same time of day. The numerical assessment can be seen in Table \ref{tab:DayandNight}

\subsection{Ablation study for Self-Attention Resolution Impact}

\begin{figure*}[ht]
  \centering
  \begin{tabular}{|c|c|c|c|}
    \hline
    \includegraphics[width=3cm, height=2cm]{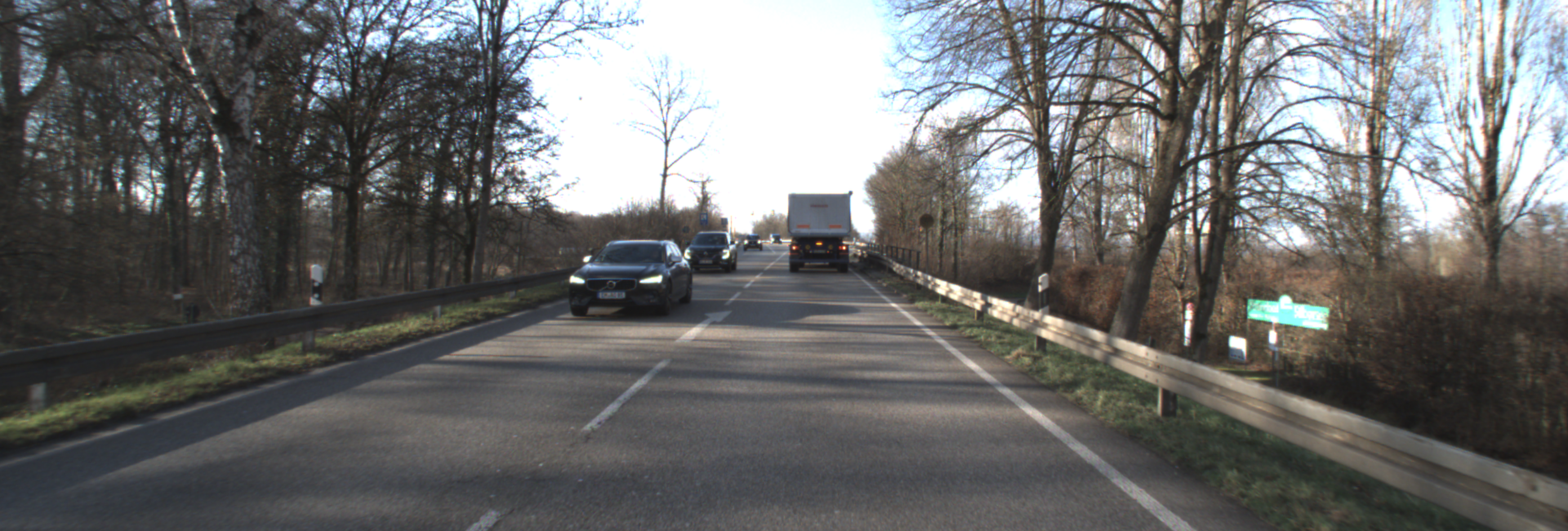} &
    \includegraphics[width=3cm, height=2cm]{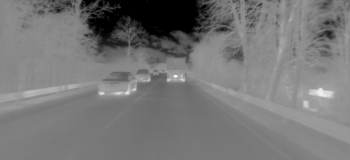} &
    \includegraphics[width=3cm, height=2cm]{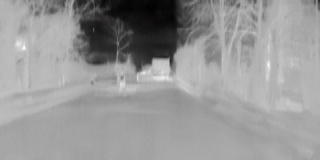} &
    \includegraphics[width=3cm, height=2cm]{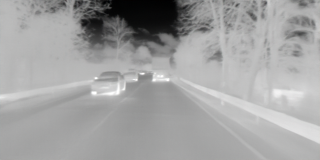} \\
    \hline
    \includegraphics[width=3cm, height=2cm]{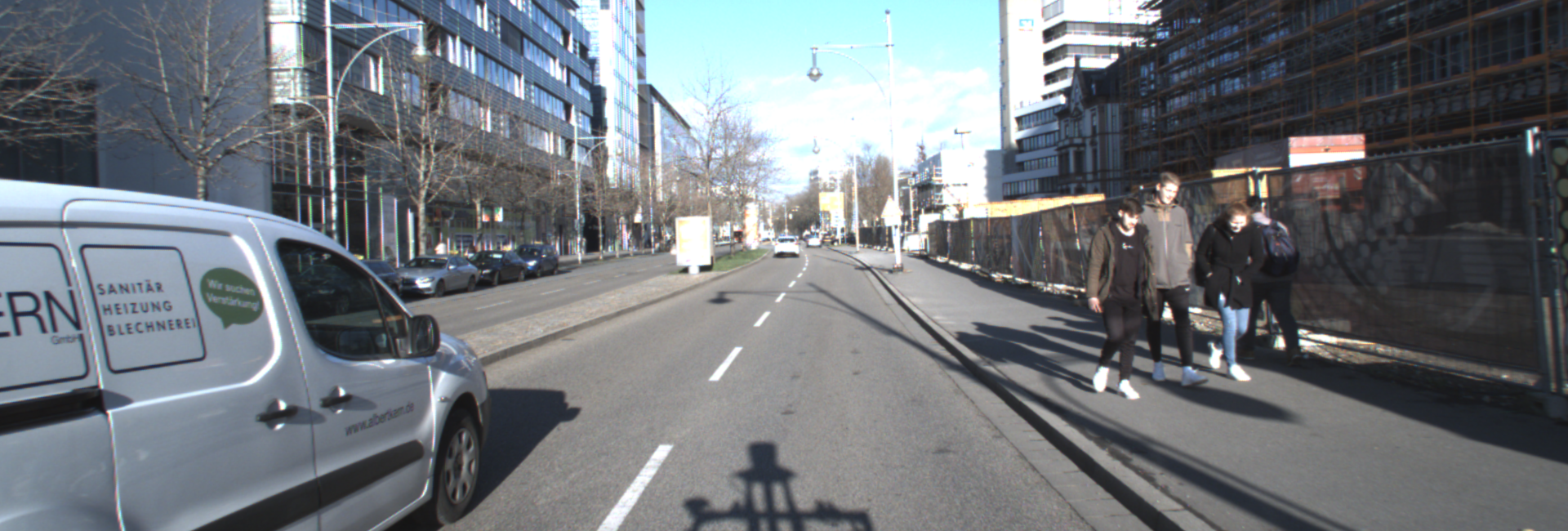} &
    \includegraphics[width=3cm, height=2cm]{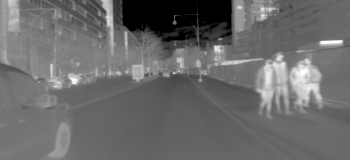} &
    \includegraphics[width=3cm, height=2cm]{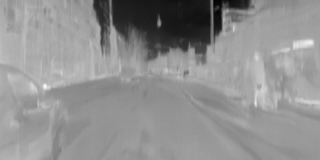} &
    \includegraphics[width=3cm, height=2cm]{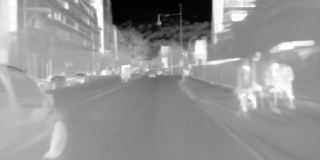} \\
    \hline
    \includegraphics[width=3cm, height=2cm]{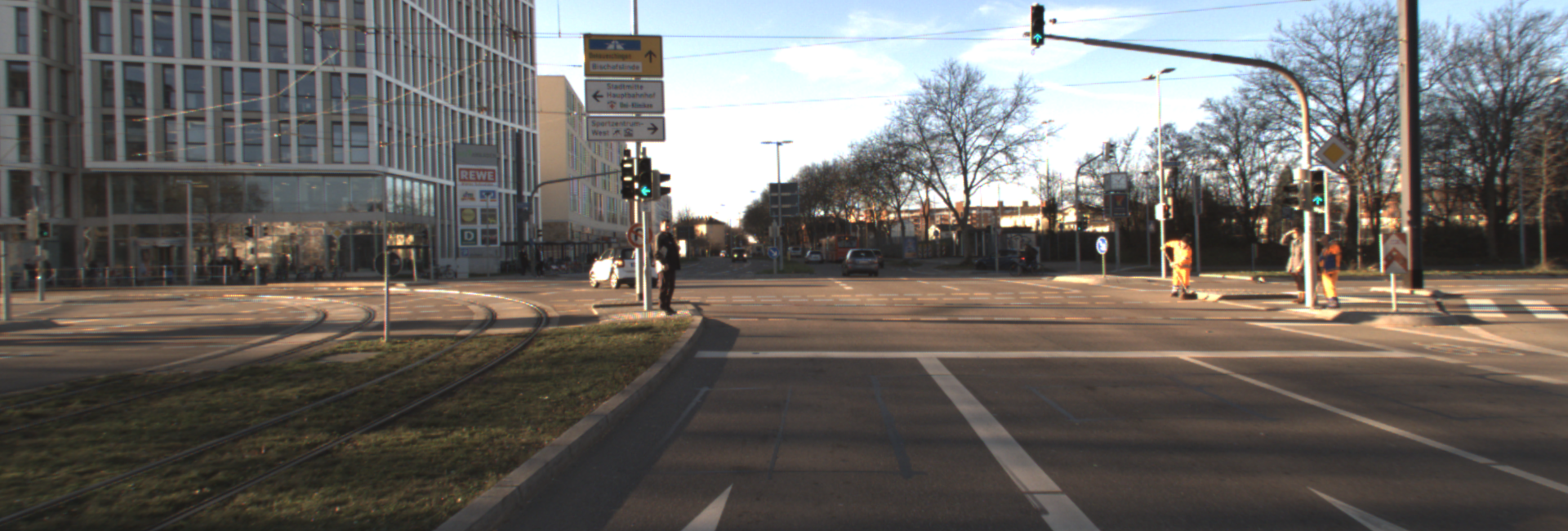} &
    \includegraphics[width=3cm, height=2cm]{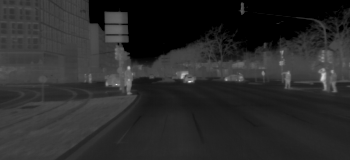} &
    \includegraphics[width=3cm, height=2cm]{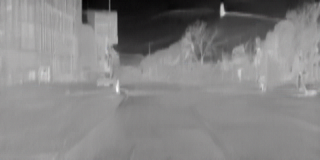} &
    \includegraphics[width=3cm, height=2cm]{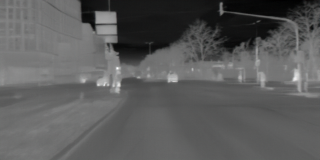} \\
    \hline
    
  \end{tabular}
  \caption{Self-attention ablation study on Freiburg Daytime Images; Column 1: RGB Images, Column 2: GT Thermal Images, Column 3: Generated Thermal Images for Self-attention Model I, Column 4: Generated Thermal Images for Self-attention Model II.}
  \label{fig:Self-attention}
\end{figure*}

We compare the qualitative and quantitative impact of increasing the resolution of self-attention in the model architecture.
The original Guided Diffusion model was trained with self-attention at the resolution levels (height/4 x width/4), (height/8 x width/8) and (height/16 x width/16) where height and width are the input image's height and width dimension, we call this Self-attention Model I. We added an additional resolution level (height/2 x width/2) with the intuition that thermal details in the images are lost significantly at the resolution of (height/4 x width/4). We call this Self-attention Model II. 
The results of our ablation study are shown in Figure \ref{fig:Self-attention} and Table \ref{tab:Self-attention}. For this study, we trained and tested on the Freiburg Daytime Images with Self-attention Models I and II.

\vspace{0.3cm}
\begin{table}[h!]
\centering
\begin{tabular}{|c|c|c|c|}

\hline
\cline{2-4}  & \multicolumn{3}{|c|}{Metrics}\\
\cline{2-4}
\cline{2-4} & PSNR $\uparrow$ & SSIM $\uparrow$ & FID $\downarrow$ \\
\hline
Self-attention Model I & 10.78 & 0.59 & 265.21 \\
\hline
Self-attention Model II & 11.14 & 0.65 & 156.53 \\
\hline

\end{tabular}
\caption{Ablation study on size of Self-attention.}
\label{tab:Self-attention}
\vspace{-0.5cm}
\end{table}

\subsection{Adaptation to Field Robotics Applications: Caltech Thermal Dataset}

The Caltech Aerial Dataset \cite{c14} comprises synchronized RGB and thermal image pairs collected over a diverse range of terrains, including rivers, coastlines, deserts, forests, and mountainous regions, captured from an aerial platform. This dataset is employed to extend our model’s applicability to field robotics, where the use of thermal imaging is crucial under low-visibility conditions, such as nighttime operations and adverse weather.

We fine-tune the day+night combined model previously trained in Section \ref{sec_exp_freiburg} using the Caltech dataset. The fine-tuned model preserves accurate recognition of previously seen categories, such as people and vehicles from the Freiburg dataset, while also adapting to new terrestrial features present in aerial imagery. Notably, the model learns to de-emphasize geometric patterns in the RGB modality in favor of thermal signatures (middle row), allowing for more reliable detection of temperature-based variations across natural landscapes. \vspace{-0.2cm}

Qualitative results of this adaptation are shown in Figure \ref{fig:CalTech}. For example, water bodies, which exhibit lower thermal intensities during the day, appear significantly darker in the synthesized images in Column 4 of all the rows. The model accurately detects people on the beach in row 3. Moreover RGB-specific geometric features like waves which are visible in RGB and absent in the thermal modality are accordingly not reproduced. Between Column 3 and 4, we see the difference between the model before being finetuned on the terrestrial data and after.

\begin{figure*}[ht]
  \centering
  \begin{tabular}{|c|c|c|c|}
    \hline
    \includegraphics[width=3cm, height=2cm]{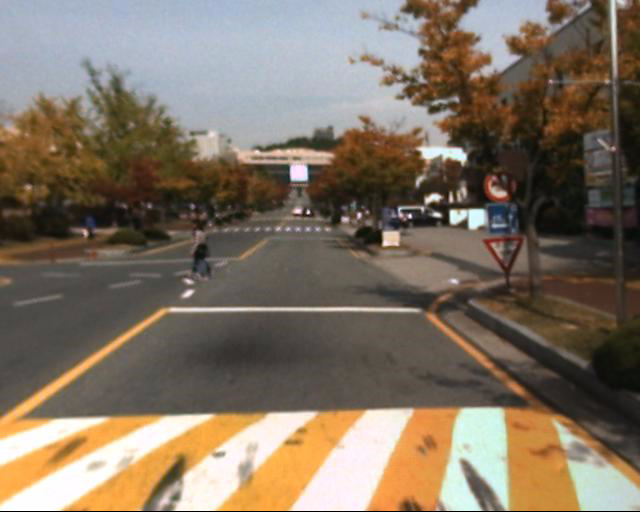} &
    \includegraphics[width=3cm, height=2cm]{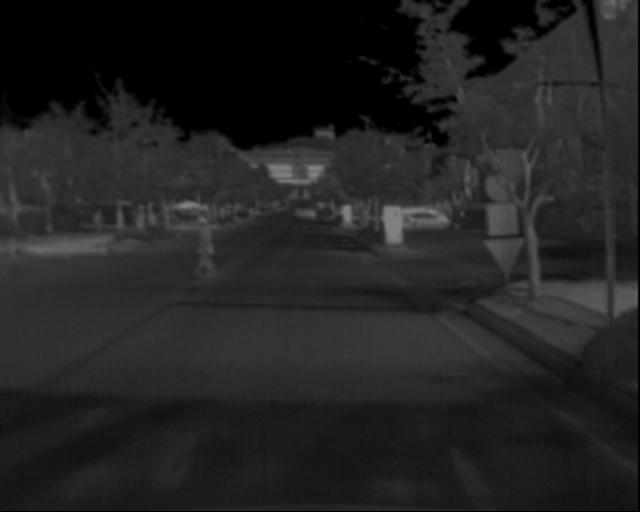} &
    \includegraphics[width=3cm, height=2cm]{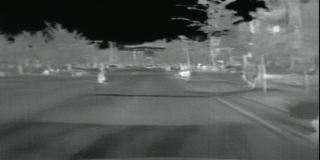} &
    \includegraphics[width=3cm, height=2cm]{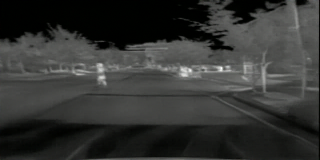} \\
    \hline
    \includegraphics[width=3cm, height=2cm]{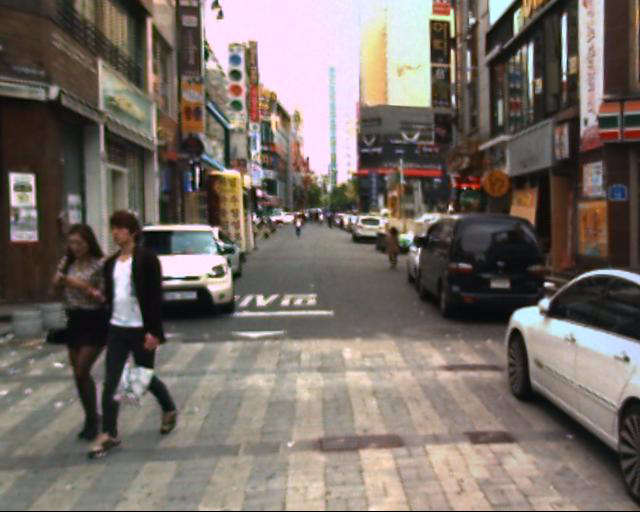} &
    \includegraphics[width=3cm, height=2cm]{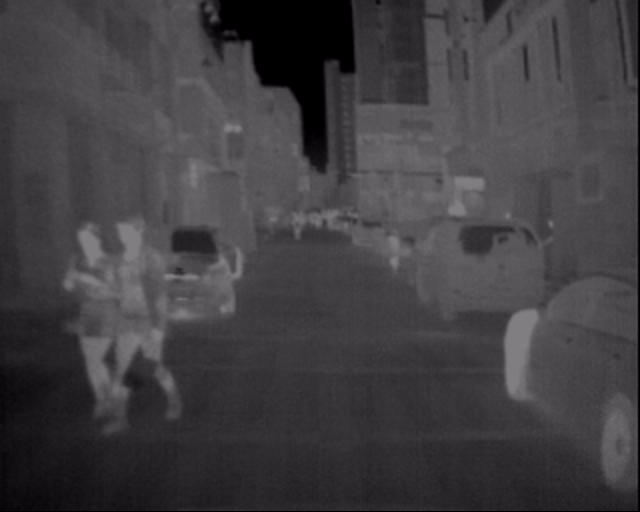} &
    \includegraphics[width=3cm, height=2cm]{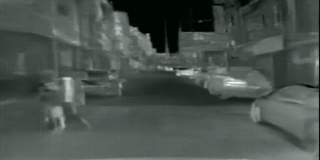} &
    \includegraphics[width=3cm, height=2cm]{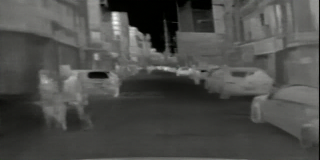} \\
    \hline
    \includegraphics[width=3cm, height=2cm]{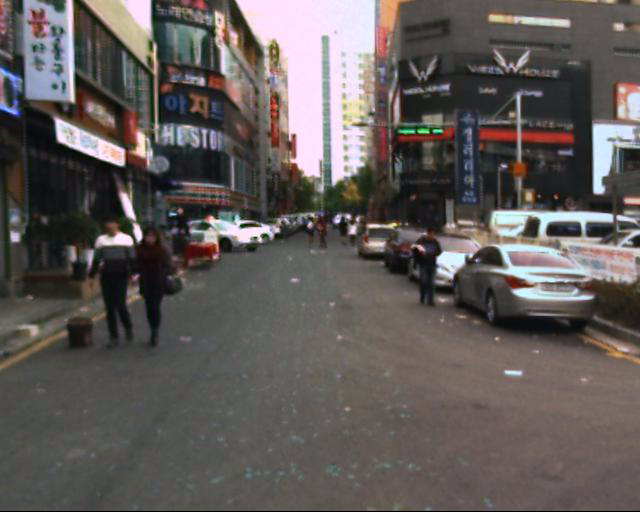} &
    \includegraphics[width=3cm, height=2cm]{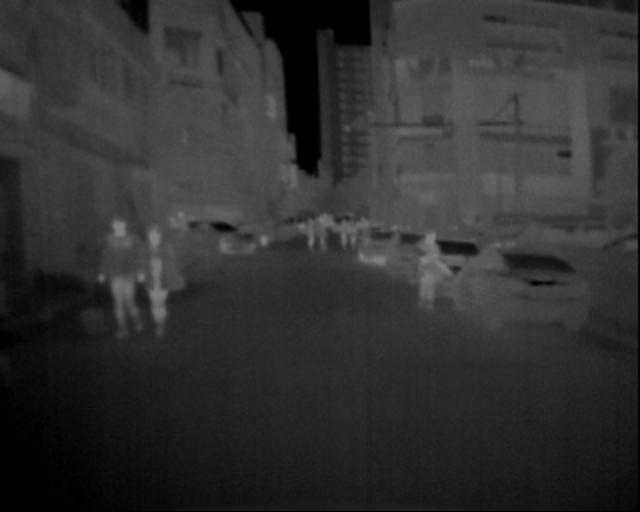} &
    \includegraphics[width=3cm, height=2cm]{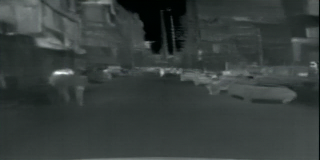} &
    \includegraphics[width=3cm, height=2cm]{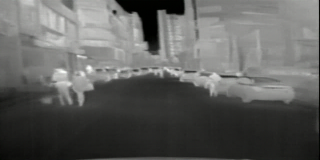} \\
    \hline
    
  \end{tabular}
  \caption{Training Data ablation on KAIST Dataset; Column 1: RGB Images, Column 2: GT Thermal Images, Column 3: Generated Thermal Images trained on KAIST Dataset only, Column 4: Generated Thermal Images pretrained on Freiburg Dataset, finetuned on KAIST Dataset.}
  \label{fig:KaistFreib}
\end{figure*}

\subsection{KAIST and FLIR Dataset Experiments}

KAIST and FLIR are two prominent thermal datasets in the field of autonomous driving. We fine-tune 
for KAIST dataset Daytime and Nighttime models separately from our Daytime and Nighttime only models in \ref{sec_exp_freiburg} while FLIR is trained on the Day + Night model due to difficulty in separating the day and night time images. We compare the performance of both models against two latent diffusion-based models, LDM \cite{ldm} and PID \cite{c16}. The qualitative results, presented in Figures \ref{fig:KAIST} and \ref{fig:FLIR}, show that our model achieves superior visual representations of scene elements—particularly pedestrians and vehicles—which are rendered with greater clarity and accuracy.

Additionally, we conduct an ablation study comparing two variants of our model: one trained from scratch on the KAIST dataset, and another pretrained on the Freiburg dataset and subsequently fine-tuned on KAIST. As shown in Figure \ref{fig:KaistFreib} and Table \ref{tab:Pre-training}, the Freiburg-pretrained model consistently outperforms the version trained solely on KAIST. We attribute this improvement to the broader and more diverse data available in the Freiburg dataset, especially its richer representation of people.

\begin{table}[h!]
\centering
\begin{tabular}{|c|c|c|c|}

\hline
\cline{2-4}  & \multicolumn{3}{|c|}{Metrics}\\
\cline{2-4}
\cline{2-4} Training Data & PSNR $\uparrow$ & SSIM $\uparrow$ & FID $\downarrow$ \\
\hline
KAIST & 14.17 & 0.50 & 62.34 \\
\hline
Freiburg + KAIST & 14.47 & 0.52 & 49.65 \\
\hline

\end{tabular}
\caption{Comparison of two models - one trained from scratch on the KAIST
dataset, and another pretrained on the Freiburg dataset and
subsequently fine-tuned on KAIST.}
\label{tab:Pre-training}
\end{table}

\section{CONCLUSIONS}

Motivated by the need to enhance autonomous driving datasets with thermal imagery that facilitates the integration of thermal cameras into autonomous navigation systems, we apply conditional diffusion models to the task of paired visual-to-thermal image translation. To maximize performance, we conducted a series of ablation studies exploring factors such as the effect of self-attention resolution, the influence of daytime versus nighttime training data, and the benefits of pre-training in datasets rich in relevant scene characteristics. Our experiments span multiple datasets, including the Freiburg Thermal Dataset, Caltech Aerial RGB-Thermal Dataset, KAIST Multispectral Pedestrian Dataset, and FLIR Thermal data set. The proposed model demonstrates strong performance in diverse environments, including urban scenes during both day and night, as well as natural outdoor settings, while preserving the important object information critical to reliable navigation.

\bibliography{IEEEabrv, refs}



\end{document}